\RequirePackage{fix-cm}
\documentclass{article}

\usepackage{iclr2027_conference,times}
\usepackage{amsmath,amsfonts,bm}

\def\eqref#1{equation~\ref{#1}}

\def\1{\bm{1}}

\DeclareMathAlphabet{\mathsfit}{\encodingdefault}{\sfdefault}{m}{sl}
\SetMathAlphabet{\mathsfit}{bold}{\encodingdefault}{\sfdefault}{bx}{n}

\newcommand{\KL}{D_{\mathrm{KL}}}

\usepackage{microtype}
\usepackage{graphicx}
\usepackage{subcaption}
\usepackage{booktabs}
\usepackage{multirow}
\usepackage{array}
\usepackage{tabularx}
\usepackage{amsmath,amssymb}
\usepackage{xcolor}
\usepackage{url}
\usepackage{hyperref}
\hypersetup{hidelinks}
\usepackage{enumitem}
\newcolumntype{Y}{>{\raggedright\arraybackslash}X}
\newcolumntype{P}[1]{>{\raggedright\arraybackslash}p{#1}}

\newcommand{\method}{\textsc{RouteVerify}}

\newcommand{\clean}{\mathrm{C}}
\newcommand{\error}{\mathrm{E}}
\newcommand{\marginm}{M}
\newcommand{\effect}{\Delta}
\newcommand{\teffect}{T}
\newcommand{\resid}{\mathrm{res}}
\newcommand{\router}{\mathrm{router}}
\newcommand{\allrouter}{\mathrm{all\mbox{-}router}}

\definecolor{cigray}{HTML}{7A7A7A}
\makeatletter
\newcommand{\ci}[1]{{\fontsize{\dimexpr\f@size pt*4/5\relax}{\dimexpr\f@size pt\relax}\selectfont\color{cigray}#1}}
\makeatother

\newcommand{\tablestyle}{\small\setlength{\tabcolsep}{5pt}}

\title{RouteVerify: Routing Interventions for\\
Step-Level Verification in\\
Mixture-of-Experts Models}

\author{Yucheng Du \\
University of Southern California \\
\texttt{yuchengd@usc.edu} \\
\And
Ryan A. Rossi \\
Adobe Research \\
\texttt{ryrossi@adobe.com} \\
\And
Xiyang Hu \\
Arizona State University \\
\texttt{xiyanghu@asu.edu}
}

\iclrpreprintcopy

\begin{document}

\maketitle
\begingroup
\renewcommand{\thefootnote}{}
\footnotetext{Code: \url{https://github.com/yucheng-du/RouteVerify}.}
\endgroup
\begin{abstract}
When prompted to judge whether a step in a given solution trace is correct, mixture-of-experts (MoE) models encode step correctness in their router logits. Replacing those logits with logits recorded on another solution trace shifts the model's judgment, yet logits from a correct versus an erroneous step rarely shift the same input differently. We test and quantify this dissociation with \textbf{RouteVerify}, a controlled probing and activation-replacement study across nine MoE checkpoints. Keeping the evaluated input fixed, we replace selected activations with those recorded on another solution trace (the \emph{donor}) at three interfaces (router logits, MoE outputs, and post-MoE residual states) across varied layers and token positions. We score the model's judgment by its log-probability preference for \texttt{INCORRECT} over \texttt{CORRECT} (the \emph{verification margin}), and measure both how strongly swapping activations between a correct and an erroneous trace shifts this margin and whether an erroneous donor shifts the same fixed input more toward \texttt{INCORRECT} than a correct donor does. Four linked results support this finding. First, the correctness signal is readable in routing: linear probes on router logits separate correct from erroneous steps in all eight checkpoints with recorded activations (AUROC 0.80--0.88), outperforming a text-only baseline. Second, routing interventions are causally potent: replacing router logits across all routed layers shifts the margin toward the donor's label in all nine checkpoints, exceeding a random perturbation of the same norm and showing that routing is not causally inert. Third, this shift rarely tracks donor correctness: on the same fixed input, erroneous and correct donors produce indistinguishable margins under single-layer routing in all nine checkpoints, and under all-layer routing an erroneous donor shifts the margin more toward \texttt{INCORRECT} in only one checkpoint, with the 95\% upper bound on this difference below 15\% of the all-layer routing shift in six checkpoints. Fourth, this near-null router result is not due to an insensitive test: under the same comparison, even single-layer replacements show this erroneous-over-correct difference in eight of nine checkpoints for residual states and five of nine for MoE outputs. Thus, readability and causal potency do not imply that a routing intervention carries the correctness signal that routing encodes.
\end{abstract}

\section{Introduction}

Step-level verifiers judge whether each intermediate step of a solution is correct, rather than only the final answer \citep{lightman2024letsverify}.
Internal representations can reveal whether an output is correct \citep{orgad2025knowmore}, and hidden-state interventions can shift step-level verification behavior \citep{zhou2026verifysteer}. Mixture-of-experts (MoE) models offer a further candidate: routing, whose logits can be probed for step correctness and edited to alter expert selection and weighting. When router logits both encode step correctness and shift the verifier's judgment when edited, it is tempting to conclude that the edit shifts the judgment by transferring that correctness signal. That inference conflates causal potency with correctness transfer, because an intervention can shift a judgment without shifting it according to the correctness of the step whose activations are inserted. Across nine MoE checkpoints, we find that router logits do encode step correctness and all-layer routing replacements do shift the verification judgment beyond random noise, yet logits from a correct versus an erroneous step rarely shift the same input differently.

\begin{figure}[t] \centering \includegraphics[width=\textwidth]{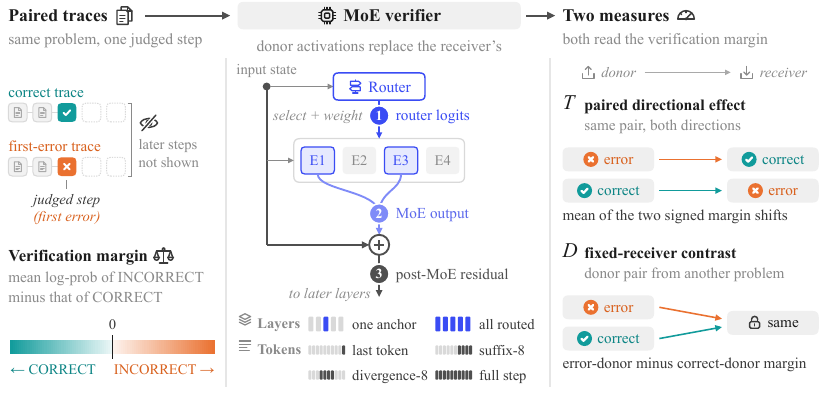} \caption{ Overview of \method{}. Activations recorded on a donor trace replace selected internal activations during the forward pass on a receiver trace, whose input text remains unchanged. We vary the intervention interface, model layers, and input token positions. \(T\) averages the shifts toward the donor's label across the two opposite-direction swaps within a correct/error pair (each compared to its unpatched baseline). \(D\) runs the same unchanged receiver input twice, replacing its internal activations once with an erroneous donor's and once with a correct donor's from another problem; positive \(D\) means the erroneous donor produces a more \texttt{INCORRECT}-favoring margin than the correct donor.\looseness=-1 } \label{fig:overview} \end{figure}

We study this question with \method{} (Figure~\ref{fig:overview}), a controlled activation-replacement study across nine MoE checkpoints. Each problem contributes a pair of solution traces: one is correct, and the other contains its first error at the evaluated step. The model judges that step as \texttt{CORRECT} or \texttt{INCORRECT}. We record internal activations while the model processes one trace, the \emph{donor}, and insert selected activations into its computation on another trace, the \emph{receiver}. Because the receiver's input text never changes, any change in its judgment comes from the inserted activations.

We quantify the judgment with a \emph{verification margin}: the difference between the model's length-normalized log probabilities of \texttt{INCORRECT} and \texttt{CORRECT}, so that larger values favor \texttt{INCORRECT}. A large shift in this margin when swapping activations between a correct and an erroneous trace does not by itself show that the model responds to the donor step's correctness. Suppose the model initially predicts \texttt{CORRECT} (negative margin) on the correct trace and \texttt{INCORRECT} (positive margin) on the erroneous trace. Any disruptive intervention that simply weakens confidence and pulls both margins toward zero will push the correct trace upward (toward \texttt{INCORRECT}) and the erroneous trace downward (toward \texttt{CORRECT}), mimicking a shift toward the donor's label without using correctness at all. We therefore separate two quantities: $T$ measures how strongly swapping activations within a pair shifts each trace's margin toward the other trace's label relative to its unpatched baseline; $D$ runs the same unchanged receiver input twice, replacing its internal activations once with those from an erroneous donor and once with those from a correct donor (from another problem), and takes the difference in the resulting margins. Because the receiver input is identical in both runs, its baseline margin and any correctness-independent disruption cancel out, leaving $D>0$ only when the erroneous donor's activations push the judgment more toward \texttt{INCORRECT} than the correct donor's.

We compare three intervention interfaces: \emph{router logits}, which score candidate experts; \emph{MoE outputs}, which carry the routed experts' combined contribution; and \emph{post-MoE residual states}, which hold the representation after the MoE update is added. We also vary which model layers and which input token positions are replaced: a single layer versus all routed layers, across four \emph{token windows} within the evaluated step (its final token, its final eight tokens, eight tokens around where the paired traces first diverge, and the whole step). Comparing these windows separates how broadly an intervention is applied (one token, eight tokens, or the full step) from where in the step it acts (the end of the step versus the point of first divergence).

Combining representation probes with $T$ (judgment-shift size) and $D$ (correctness dependence on a fixed receiver), four linked results support this claim:
\begin{itemize}[leftmargin=*,itemsep=2pt,topsep=3pt]
  \item \textbf{Router logits are decodable for step correctness.} Out-of-fold
  linear probes on router logits separate correct from erroneous steps in all
  eight checkpoints with recorded activations (AUROC 0.80--0.88), outperforming
  a text-only baseline on the evaluated step (Section~\ref{sec:probes}).
  \item \textbf{Routing interventions are causally potent.} Replacing router
  logits across all routed layers yields positive $T$ in all nine checkpoints
  and exceeds a random perturbation of the same norm, showing that routing is
  not causally inert. Although single-layer residual replacements produce
  larger $T$ than single-layer router or MoE-output replacements in all 36
  checkpoint-by-window comparisons, all-layer routing matches or exceeds
  single-layer residual replacement in three checkpoints, and moving an
  eight-token window from the end of the step to the first divergence narrows
  the residual-versus-routing gap in all nine checkpoints
  (Sections~\ref{sec:local} and~\ref{sec:windows}).
  \item \textbf{The routing shift rarely carries donor correctness.} Holding
  the receiver fixed, router $D$ is unresolved (its 95\% interval includes
  zero) at a single layer in all nine checkpoints and reliably positive under
  all-layer routing in only one checkpoint (GPT-OSS-20B). In six checkpoints,
  the upper 95\% bound of all-layer router $D$ is below 15\% of the all-layer
  routing shift $T$ (Section~\ref{sec:router-d}).
  \item \textbf{This near-null router result is not an insensitive test.} Under
  the same fixed-receiver comparison, even single-layer replacements yield
  reliably positive $D$ on both correct and erroneous receivers in eight of
  nine checkpoints for post-MoE residual states and in five of nine for MoE
  outputs, confirming that an unresolved router $D$ reflects the interface
  rather than an insensitive measurement (Section~\ref{sec:donor-role}).
\end{itemize}
Taken together, these four results show that decodability and causal potency do not imply correctness transfer: router logits encode step correctness and editing them moves the verification margin, yet routing interventions rarely carry correctness into the model's judgment.

\section{Intervention Design}
\label{sec:method}

\subsection{Verification task and score}

Our evaluation set is derived from PRMBench, a benchmark with step-level
annotations for evaluating fine-grained error detection in reasoning
traces \citep{song2025prmbench}. From 580 complete pairs of solution traces split 50/50 by error type, all
experiments in this paper use one 290-pair half (the discovery split), while
the remaining 290 pairs are held out untouched
(Appendix~\ref{app:model-protocols}). In each pair, one trace is labeled
correct and the other contains its first annotated error at the evaluated
step. The model receives the problem, the solution trace
through this step, and a fixed instruction to judge the step as
\texttt{CORRECT} or \texttt{INCORRECT}. Later steps and analysis metadata
are not provided. We write $\clean$ and $\error$ for the correct and
erroneous members of a pair, respectively.

Rather than recording only which label the model prefers, we measure how
strongly it prefers one over the other with a continuous
\emph{verification margin}:
\begin{equation}
\marginm(x)
=
|y_{\mathrm I}|^{-1}\log p(y_{\mathrm I}\mid x)
-
|y_{\mathrm C}|^{-1}\log p(y_{\mathrm C}\mid x),
\label{eq:margin}
\end{equation}
where $y_{\mathrm I}$ and $y_{\mathrm C}$ are the model-specific token
sequences for \texttt{INCORRECT} and \texttt{CORRECT}, and $|y|$ denotes
the number of tokens in a sequence. Positive values favor
\texttt{INCORRECT}; negative values favor \texttt{CORRECT}. An increase in
the margin therefore indicates a shift toward an \texttt{INCORRECT}
judgment, although it need not change the preferred label. Per-token averaging prevents differences in label token length from biasing the
margin; total-logprob margins serve as a sensitivity analysis.

\subsection{Activation replacement and two intervention measures}

\textbf{Donors, receivers, and replacement.}
We call the example supplying activations the \emph{donor} and
the example whose activations are replaced the \emph{receiver}. The
receiver retains its original problem, solution text, and verification
instruction. Only the selected internal activations are replaced; the
model then continues its computation and scores the two judgment labels.

For an intervention setting $s$, let $h_s^r$ and $h_s^d$ denote the
corresponding receiver and donor activations. We replace the receiver
activations according to
\begin{equation}
\widetilde h_s^r
=
h_s^r+\alpha(h_s^d-h_s^r),
\label{eq:replacement}
\end{equation}
where $\alpha$ controls the replacement strength. All reported experiments
use $\alpha=1$, which is exact replacement. The setting $s$ specifies
the intervention interface and its layer and token coverage. Let
$x_{r\leftarrow d,s}$ denote the receiver evaluated under this intervention.
Its margin change is
\begin{equation}
\effect_{d\rightarrow r,s}
=
\marginm(x_{r\leftarrow d,s})-\marginm(x_r).
\label{eq:margin-change}
\end{equation}

\textbf{Paired directional effect, $T$.}
For each correct/error pair, we perform two replacements: activations from
the erroneous trace are inserted into the correct receiver, and activations
from the correct trace are inserted into the erroneous receiver. We
summarize these two changes as
\begin{equation}
\teffect_s
=
\tfrac12
\left(
\effect_{\error\rightarrow\clean,s}
-
\effect_{\clean\rightarrow\error,s}
\right).
\label{eq:t}
\end{equation}
The signs are chosen so that an increase in the correct receiver's margin
and a decrease in the erroneous receiver's margin both contribute positively.
Thus, positive $T$ indicates an average shift toward the donor step's label
across the two replacements; it does not require both replacements to move
in that direction.\looseness=-1

However, $T$ can also be positive when an intervention simply reduces the
original separation between the two receivers:
\begin{equation}
2\teffect_s
=
[\marginm(x_\error)-\marginm(x_\clean)]
-
[\marginm(x_{\error\leftarrow\clean,s})
-
\marginm(x_{\clean\leftarrow\error,s})].
\label{eq:gap-compression}
\end{equation}
For example, an intervention that makes both receivers' margins approach
zero can yield positive $T$ when their original margins favor their
respective labels. Consequently, $T$ measures a signed paired intervention
effect, but does not by itself establish that the effect depends on
whether the donor step is correct or erroneous.

\textbf{Fixed-receiver donor contrast, $D$.}
To test this dependence, we compare correct and erroneous donors on the
same receiver. For each receiver pair $i$, we select an external matched
pair $j$ from another problem. We then evaluate each receiver
$r\in\{\clean_i,\error_i\}$ twice at the same intervention setting:
once with activations from the correct donor and once with activations
from the erroneous donor. We define
\looseness=-1
\begin{equation}
D_{i,r,s}
=
\marginm(x_{r\leftarrow\error_j,s})
-
\marginm(x_{r\leftarrow\clean_j,s}),
\qquad
D_{i,\mathrm{common},s}
=
\tfrac12
\left(
D_{i,\clean_i,s}+D_{i,\error_i,s}
\right).
\label{eq:d}
\end{equation}
Positive $D$ means that, for the same receiver, the erroneous donor produces
a more \texttt{INCORRECT}-favoring margin than the correct donor, and
$D_{\mathrm{common}}$ averages this contrast over the two receiver types. Because
the receiver input text is identical in both runs, its baseline margin and
any correctness-independent disruption cancel from the comparison.
This contrast compares matched donor steps of opposite correctness, though it
cannot isolate correctness from every correlated activation difference. Appendix~\ref{app:surface-routing} further tests it against a
same-label placebo ordered by a lexical (TF-IDF) proxy.

\begingroup
\parfillskip=0pt plus .5\linewidth
\finalhyphendemerits=100000
$T$ and $D$ therefore answer different questions. $T$ summarizes signed
changes from the two cross-role replacements within a correct/error pair,
whereas $D$ compares two donor roles on an unchanged receiver. Because they
use different donor--receiver comparisons, neither $D/T$ nor $T-D$ has a
decomposition interpretation.\par
\endgroup

\subsection{Interfaces, layer coverage, and token windows}
\label{sec:interfaces}

\textbf{Intervention interfaces.}
We replace activations at three points in an MoE block. A
\emph{router-logit} intervention replaces the scores assigned to candidate
experts before expert selection. The model then uses its native expert
selection, weighting, and dispatch procedures. A \emph{MoE-output}
intervention replaces the routed experts' combined output before it is
added to the residual stream. A \emph{post-MoE residual} intervention
replaces the resulting residual state, which can already contain
information from attention, experts, and earlier layers.

Where an architecture includes a shared-expert or parallel dense branch,
that branch remains receiver-derived under the MoE-output intervention
(Section~\ref{sec:strata}). Although all three interfaces are evaluated
using the same verification margin, they replace different amounts of
computation. A residual replacement swaps more of the computation than
a router replacement, so a larger residual effect does not show that routing
is unimportant. The like-for-like comparison is router logits against MoE outputs:
both target the routed update, the former indirectly through expert selection
and weighting and the latter directly. The three interventions
are not additive stages of a single mediation decomposition.

\textbf{Layer and token coverage.}
For local comparisons, we use one fixed, model-specific layer, referred to
as the \emph{anchor layer}. At this layer, we compare all three interfaces
under four \emph{token windows}, each specifying the positions within the
evaluated reasoning step at which activations are replaced: the last token;
the final eight tokens (\emph{suffix-8}); eight tokens around the first
token at which the paired prefixes diverge (\emph{divergence-8}); and the
full step.

We additionally apply a joint \emph{all-router} intervention, replacing
router logits at every routed layer under the same four token windows.
Comparing suffix-8 with divergence-8 holds the number of intervened tokens
fixed while changing their position. It does not, however, equalize the
size of the activation changes or the computation preceding those
positions. Exact window construction and donor--receiver token alignment
are given in Appendix~\ref{app:nine-model-protocols}.

We evaluate $D$ both locally, at one layer and the final step token, and
in a separate experiment that replaces router logits across all routed
layers and the full step (Appendix~\ref{app:wide-d}). We also evaluate an
\emph{all-layer MoE-output} intervention at the last token; this experiment
is reported with $T$ only.

\textbf{Representation probes.}
To test whether step correctness is linearly readable before intervening, we
fit out-of-fold L2-regularized logistic probes on recorded last-token
activations (post-MoE residual states, router inputs, router logits, and
expert-aligned sparse weights), using five folds grouped by source problem so
that pairs from the same problem never span training and evaluation folds. We
compare router-logit AUROC against a fixed word unigram/bigram TF-IDF logistic
baseline on the evaluated step trained on the same folds
(Appendix~\ref{app:access}).

\subsection{Controls and uncertainty}

Self-patches, which replace receiver activations with their own recorded
values, and unpatched reruns verify that the intervention machinery
preserves the receiver when no cross-example replacement is made. Matched
external donors come from other problems, prioritizing agreement in error
type and proximity of the error position, and remain fixed during analysis.
Norm-random controls replace the donor--receiver activation difference
with a Gaussian direction having the same $\ell_2$ norm. For $T$, each matched external donor has the same correctness label as the
paired donor it replaces (erroneous for $\clean$, correct for $\error$); for
$D$, both the correct and erroneous members of the matched external pair are
evaluated on each fixed receiver.

Pairs receive equal weight. To account for dependence among pairs
associated with the same source problem, we resample the $K=235$
source-problem components and retain all pairs belonging to each sampled
component. Last-token, full-step, $D$, and all-layer-output analyses use
all $N=290$ pairs. Eight-token and direct-contrast analyses use the pairs
eligible in every compared condition, with unavailable windows treated as
missing rather than zero.

Intervals are two-sided 95\% pointwise percentile intervals from 50,000
bootstrap resamples generated with PCG64. We call an estimate
\emph{resolved} when its interval excludes zero; an unresolved interval
does not establish equivalence. Direct differences are formed within pair
before resampling, and reused matched donors are conditioned on rather than
resampled independently. Exact seeds and
version-specific analysis details are reported in
Appendices~\ref{app:nine-model-protocols} and~\ref{app:wide-d}.

The checkpoints are fixed and the pair sample was assembled
retrospectively, so these intervals do not describe a population of
models. We apply no multiplicity correction; post hoc
receiver-role and GPT-checkpoint contrasts are reported as secondary
analyses.

\section{Models and Intervention Boundaries}
\label{sec:strata}

We evaluate nine MoE checkpoints: Qwen3.6-35B-A3B, Qwen3-30B-A3B,
GLM-4.7-Flash, Nemotron-3.5 Lightning 30B-A3B-BF16, Nemotron-3 Nano
30B-A3B, Gemma 4 26B-A4B, OLMoE-1B-7B, and GPT-OSS-20B/120B
\citep{yang2025qwen3,muennighoff2025olmoe,openai2025gptoss,zai2026glm47flash}.\looseness=-1

\begingroup
\parfillskip=0pt plus .5\linewidth
\finalhyphendemerits=100000
Each checkpoint uses a fixed architecture-specific anchor layer for local
interventions, while all-layer interventions cover every routed block in
that checkpoint. Appendix~\ref{app:model-protocols} reports model sources,
anchor indices, and routed-layer counts (Table~\ref{tab:abc-anchors}).
Qwen3 is the only checkpoint for which the local $D$ and token-window
experiments use different anchors: L36 and L24, respectively. Its probe
protocol is described separately (Appendix~\ref{app:access}).\par
\endgroup

The MoE-output intervention follows each architecture's native computation
boundaries. For GLM, Qwen3.6, and the Nemotron models, we replace the routed
experts' contribution while retaining the receiver's shared-expert branch.
For Gemma, we replace the normalized routed branch while retaining the
receiver's parallel dense branch. For Qwen3, OLMoE, and GPT-OSS, the
intervention replaces the native MoE/MLP-module output. Exact adapter
definitions and execution checks are reported in
Appendices~\ref{app:architecture-boundaries} and~\ref{app:controls-full}, respectively.

\section{Results}
\label{sec:prm-results}

Every checkpoint verifies the paired PRMBench steps above chance before any
intervention, though with substantially different quality. Reading
$\marginm>0$ as an \texttt{INCORRECT} judgment, unpatched accuracy ranges
from $0.57\;[0.55,0.59]$ in OLMoE to $0.79\;[0.76,0.82]$ in Gemma, and AUROC
from $0.60$ to $0.87$, against a chance level of $0.5$ on this balanced
assay (Appendix~\ref{app:verification-metrics}). The subsections below
follow the four linked results in order.\looseness=-1

\subsection{Router logits are decodable for step correctness}
\label{sec:probes}

Routing carries a strong, linearly readable correctness signal across every
probed checkpoint. Out-of-fold L2-logistic probes on last-token router logits,
with folds grouped by source problem, separate correct from erroneous steps
well above the $0.50$ chance level in all eight checkpoints with recorded
activations, reaching AUROC from $0.80$ in GPT-OSS-20B to $0.88$ in Qwen3
(Table~\ref{tab:interface-access} in Appendix~\ref{app:access}; OLMoE has no
recorded activations). Across representation sites, router logits consistently
sit just below post-MoE residual states ($0.83$--$0.91$) and router inputs
($0.84$--$0.92$), and well above sparse expert-aligned routing weights
($0.69$--$0.79$).

This decodability is not merely a bag-of-words artifact of the evaluated step
text. On the same five folds, router logits outperform a fixed boundary-step
TF-IDF logistic baseline (AUROC $0.74$) by $+0.06$ to $+0.12$ across all five
checkpoints sharing that boundary protocol, with every paired 95\% interval
excluding zero (Table~\ref{tab:text-baseline} in Appendix~\ref{app:access}).
Decodability alone, however, does not show that the model uses routing to
carry correctness into its judgment. Sections~\ref{sec:local}--\ref{sec:windows}
test whether replacing routing causally shifts the verification judgment, and
Sections~\ref{sec:router-d}--\ref{sec:donor-role} test whether that shift
depends on donor correctness.

\subsection{At a single layer, residual replacements move the judgment most}
\label{sec:local}

\begin{figure}[!t]
\centering
\includegraphics[width=\textwidth]{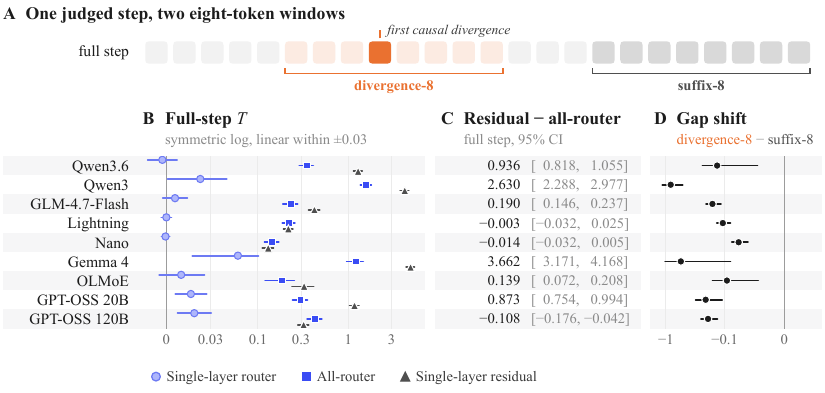}
\caption{
\textbf{Interface and coverage effects on the verification margin.}
\textbf{A}: positions of the divergence-8 and suffix-8 windows within an
evaluated step.
\textbf{B}: full-step $T$ for single-layer router, all-router, and residual
interventions.
\textbf{C}: the full-step residual-minus-all-router difference.
\textbf{D}: the change in this difference from suffix-8 to divergence-8;
negative values indicate a smaller residual-versus-routing gap near the first
divergence. Error bars are pointwise 95\% source-component percentile
intervals from 50,000 resamples; direct contrasts use pairs eligible in both
conditions.
Panels \textbf{B} and \textbf{D} use symmetric-log axes.\looseness=-1}
\label{fig:abc-windows}
\end{figure}

At a fixed layer, routing is the weakest of the three interfaces. Across
nine checkpoints and four token windows, post-MoE residual replacements
produce larger $T$ than both router-logit and MoE-output replacements: all
36 residual-minus-router and all 36 residual-minus-output direct intervals
are positive. Effect sizes vary widely across checkpoints, with full-step
residual $T$ ranging from 0.13 in Nano to 4.87 in Gemma
(Figure~\ref{fig:abc-windows}B); these values compare interfaces within a
checkpoint and do not rank model quality.

Normalizing by the baseline margin gap of the same pairs,
$R=\mathrm{mean}(\teffect)/\mathrm{mean}(g)$ with
$g=\marginm(x_\error)-\marginm(x_\clean)$, places full-step residual effects
between $0.24\;[0.22,0.26]$ in GPT-OSS-120B and $0.90\;[0.88,0.93]$ in
GPT-OSS-20B. In GPT-OSS-20B, Qwen3, Qwen3.6, and Gemma, the $R$ interval lies
above $0.5$: since $R=0.5$ eliminates the mean paired gap of
Equation~\ref{eq:gap-compression}, the residual replacement reverses the gap
on average, leaving patched accuracy of $0.27$ to $0.37$
(Appendix~\ref{app:verification-metrics}).

\subsection{All-layer routing swaps are causally potent beyond random controls}
\label{sec:windows}
\label{sec:controls}

Replacing routing across all routed layers shows that the router is not
causally inert. Full-step all-router $T$ is positive in all nine checkpoints
and exceeds the corresponding single-router effect
(Figure~\ref{fig:abc-windows}B). At full
step, the residual-minus-all-router gap remains positive in six checkpoints,
is unresolved in Lightning and Nano, and reverses in GPT-OSS-120B
($-0.11\;[-0.18,-0.04]$; Figure~\ref{fig:abc-windows}C). At divergence-8,
all-router also exceeds residual in both Nemotron checkpoints and
GPT-OSS-120B. Expanded across layers and tokens, routing can thus match or
exceed a local residual replacement.\looseness=-1

Token position shapes this comparison more consistently than token count.
Expanding coverage from the last token to the full step produces no
consistent trend in the residual-versus-routing gap, but with eight tokens
fixed, moving the window from suffix-8 to divergence-8 reduces
$G_w=T_{\resid,w}-T_{\allrouter,w}$ in all nine checkpoints, with every
interval for $G_{\mathrm{divergence8}}-G_{\mathrm{suffix8}}$ below zero
(Figure~\ref{fig:abc-windows}D). The effect is relative: in GPT-OSS-120B the
narrower gap near divergence comes mainly from residual sensitivity to
position rather than from a larger routing response
(Appendix~\ref{app:nine-model-boundaries}).

Perturbation size does not explain these routing effects. Full-step
all-router $T$ exceeds the norm-random control in all nine checkpoints,
although OLMoE's interval barely excludes zero (lower bound $0.0003$;
Table~\ref{tab:abc-window-router-controls}). Replacing MoE outputs jointly at
all routed layers likewise gives
positive last-token $T$ in all nine checkpoints, from $0.10$ in Qwen3.6 to
$1.13$ in Gemma, with every paired-minus-norm-random interval positive
(Table~\ref{tab:abc-bundle}).

Neither effect, however, is specific to the paired trace. Matched external
donors from other problems, chosen to share error type and first-error
position, produce larger all-router $T$ than paired donors in eight
checkpoints (OLMoE unresolved), and every all-layer MoE-output
paired-minus-matched interval includes zero. Because matched donors can
differ in magnitude, context, and structure, a larger matched-donor effect is
not evidence of more correctness information. Verdict flips agree: full-step
all-router replacement flips $0.06$ to $0.23$ of verdicts with paired
donors, against $0.03$ to $0.11$ under the norm-random control and $0.07$ to
$0.35$ with matched donors, which exceed paired donors in all nine
checkpoints (Appendix~\ref{app:verification-metrics}). These controls show
that routing replacements move the judgment beyond what perturbation size
explains, but not whether the movement follows the donor step's correctness,
which is the question $D$ addresses.

\subsection{Routing effects rarely depend on donor correctness}
\label{sec:router-d}
\label{sec:wide-router-d}

\begin{figure}[t]
\centering
\includegraphics[width=\linewidth]{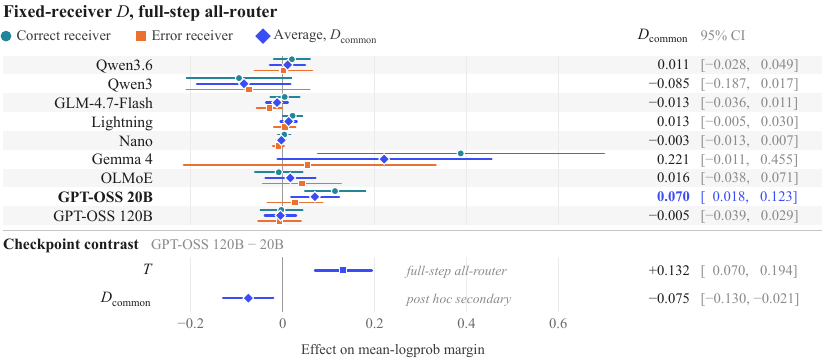}
\caption{
\textbf{Fixed-receiver effects for full-step all-router interventions.}
Positive $D$ means that, on the same receiver, activations from an erroneous
step produce a more \texttt{INCORRECT}-favoring margin than activations from a
correct step. Circles and squares show the two receiver types, and diamonds show
their average $D_{\mathrm{common}}$; the only $D_{\mathrm{common}}$ interval excluding zero
is set in bold. Error bars are pointwise 95\%
source-component percentile intervals ($N=290$, $K=235$, 50,000 resamples).
The lower strip reports direct GPT-OSS checkpoint contrasts for $T$ and
$D_{\mathrm{common}}$.}
\label{fig:wide-d-roles}
\end{figure}

Holding the receiver fixed, correct and erroneous donors seldom produce
different routing effects, at one layer or across all routed layers. At one
layer and the last token, every single-router $D_{\mathrm{common}}$ interval
includes zero (Table~\ref{tab:abc-d}). Extending the contrast across all
routed layers and the full step does not produce a consistent positive
effect either (Figure~\ref{fig:wide-d-roles}). Only GPT-OSS-20B has a
resolved positive $D_{\mathrm{common}}$, $0.07\;[0.02,0.12]$, carried by its
correct receivers ($0.11\;[0.05,0.18]$, against $0.03\;[-0.03,0.09]$ on error
receivers; direct difference $0.09\;[0.02,0.16]$). Gemma has a positive
correct-receiver effect, $0.39\;[0.08,0.70]$, but an unresolved
$D_{\mathrm{common}}$ of $0.22\;[-0.01,0.45]$. Relative to all-router $T$, the
upper interval bound of $D_{\mathrm{common}}$ is below 15\% in six checkpoints
(Appendix~\ref{app:relative-effects}). Routing diagnostics confirm that this
near-zero contrast is not a trivial failure to perturb routing: single-layer
router replacements already alter top-$k$ expert selection in 35\%--93\% of
paired cases (45\%--100\% in $D$ runs), and erroneous and correct external
donors produce similar average router KL (ratio 0.90--1.09;
Appendix~\ref{app:routing-manipulation}).

\begin{table}[!htbp]
\centering
\caption{
Fixed-receiver $D$ at the last token. Positive values mean that, on the same
receiver, activations from an erroneous step produce a more
\texttt{INCORRECT}-favoring margin than activations from a correct step.
Residual $D$ is positive on both receiver types in eight checkpoints; Qwen3.6
($\dagger$) is the exception. A substantial part of OLMoE's residual $D$ is
associated with surface content (Appendix~\ref{app:surface-routing}). Router
and output columns report $D_{\mathrm{common}}$, while residual columns retain the
two receiver types. All panels use $N=290$ pairs and $K=235$ source components
with pointwise 95\% percentile intervals.}
\begingroup
\renewcommand{\arraystretch}{1.0}%
\tablestyle
\setlength{\tabcolsep}{2pt}%
\begin{tabular}{@{}lcccc@{}}%
\toprule
Model & \begin{tabular}[t]{@{}c@{}}Residual\\\(D_{\mathrm{correct}}\)\end{tabular} & \begin{tabular}[t]{@{}c@{}}Residual\\\(D_{\mathrm{error}}\)\end{tabular} & \begin{tabular}[t]{@{}c@{}}Router\\\(D_{\mathrm{common}}\)\end{tabular} & \begin{tabular}[t]{@{}c@{}}Output\\\(D_{\mathrm{common}}\)\end{tabular} \\%
\midrule
Qwen3.6\(^{\dagger}\) & \({-}.0096\)\hspace{0.25em}\ci{\([{-}.0308, \phantom{1}.0130]\)} & \(.0356\)\hspace{0.25em}\ci{\([.0101, .0625]\)} & \({-}.0010\)\hspace{0.25em}\ci{\([{-}.0085, .0062]\)} & \({-}.0053\)\hspace{0.25em}\ci{\([{-}.0120, .0010]\)}\\[0.5pt]%
Qwen3 & \(\phantom{{-}}.7282\)\hspace{0.25em}\ci{\([\phantom{{-}}.6313, \phantom{1}.8269]\)} & \(.5921\)\hspace{0.25em}\ci{\([.4989, .6879]\)} & \({-}.0073\)\hspace{0.25em}\ci{\([{-}.0151, .0006]\)} & \(\phantom{{-}}.0519\)\hspace{0.25em}\ci{\([\phantom{{-}}.0339, .0698]\)}\\[0.5pt]%
GLM-4.7-Flash & \(\phantom{{-}}.1989\)\hspace{0.25em}\ci{\([\phantom{{-}}.1594, \phantom{1}.2404]\)} & \(.1371\)\hspace{0.25em}\ci{\([.1077, .1673]\)} & \(\phantom{{-}}.0009\)\hspace{0.25em}\ci{\([{-}.0062, .0080]\)} & \(\phantom{{-}}.0642\)\hspace{0.25em}\ci{\([\phantom{{-}}.0475, .0806]\)}\\[0.5pt]%
Lightning & \(\phantom{{-}}.1492\)\hspace{0.25em}\ci{\([\phantom{{-}}.1285, \phantom{1}.1704]\)} & \(.1125\)\hspace{0.25em}\ci{\([.0953, .1304]\)} & \({-}.0001\)\hspace{0.25em}\ci{\([{-}.0035, .0034]\)} & \({-}.0003\)\hspace{0.25em}\ci{\([{-}.0041, .0036]\)}\\[0.5pt]%
Nano & \(\phantom{{-}}.0793\)\hspace{0.25em}\ci{\([\phantom{{-}}.0680, \phantom{1}.0907]\)} & \(.0624\)\hspace{0.25em}\ci{\([.0525, .0726]\)} & \(\phantom{{-}}.0005\)\hspace{0.25em}\ci{\([{-}.0017, .0026]\)} & \(\phantom{{-}}.0004\)\hspace{0.25em}\ci{\([{-}.0024, .0032]\)}\\[0.5pt]%
Gemma 4 & \(\phantom{{-}}\llap{\(1\)}.1705\)\hspace{0.25em}\ci{\([\phantom{{-}}.8775, 1.4791]\)} & \(.2586\)\hspace{0.25em}\ci{\([.1150, .4163]\)} & \({-}.0061\)\hspace{0.25em}\ci{\([{-}.0208, .0087]\)} & \(\phantom{{-}}.1188\)\hspace{0.25em}\ci{\([\phantom{{-}}.0681, .1793]\)}\\[0.5pt]%
OLMoE & \(\phantom{{-}}.1368\)\hspace{0.25em}\ci{\([\phantom{{-}}.0859, \phantom{1}.1925]\)} & \(.1586\)\hspace{0.25em}\ci{\([.1049, .2169]\)} & \(\phantom{{-}}.0050\)\hspace{0.25em}\ci{\([{-}.0011, .0110]\)} & \(\phantom{{-}}.0349\)\hspace{0.25em}\ci{\([\phantom{{-}}.0211, .0487]\)}\\[0.5pt]%
GPT-OSS-20B & \(\phantom{{-}}.1352\)\hspace{0.25em}\ci{\([\phantom{{-}}.1016, \phantom{1}.1696]\)} & \(.1019\)\hspace{0.25em}\ci{\([.0726, .1323]\)} & \({-}.0017\)\hspace{0.25em}\ci{\([{-}.0060, .0027]\)} & \({-}.0014\)\hspace{0.25em}\ci{\([{-}.0108, .0079]\)}\\[0.5pt]%
GPT-OSS-120B & \(\phantom{{-}}.2262\)\hspace{0.25em}\ci{\([\phantom{{-}}.1895, \phantom{1}.2645]\)} & \(.1654\)\hspace{0.25em}\ci{\([.1342, .1977]\)} & \(\phantom{{-}}.0061\)\hspace{0.25em}\ci{\([{-}.0025, .0150]\)} & \(\phantom{{-}}.0436\)\hspace{0.25em}\ci{\([\phantom{{-}}.0312, .0564]\)}\\[0.5pt]%
\bottomrule
\end{tabular}%
\endgroup

\label{tab:abc-d}
\end{table}

The two GPT-OSS checkpoints show that response size and correctness
dependence can order checkpoints differently. GPT-OSS-120B has the larger
full-step all-router $T$ (120B minus 20B: $0.13\;[0.07,0.19]$), yet
GPT-OSS-20B has the larger $D_{\mathrm{common}}$ (20B minus 120B:
$0.08\;[0.02,0.13]$), and GPT-OSS-120B's own $D_{\mathrm{common}}$ is near zero
($-0.01\;[-0.04,0.03]$). A larger routing-induced judgment change therefore
need not imply stronger dependence on donor correctness. These contrasts do
not isolate model scale, because training, routed depth, and expert count
also differ (Appendix~\ref{app:wide-d}).

\subsection{The same $D$ test detects correctness in residual and MoE-output states}
\label{sec:donor-role}

The same fixed-receiver contrast detects correctness dependence at the
residual interface. At one layer and the last token, residual $D$ is positive
on both correct and error receivers in eight of nine checkpoints
(Table~\ref{tab:abc-d}), and residual-minus-router and residual-minus-output
contrasts in $D_{\mathrm{common}}$ are positive in the same eight. MoE-output $D$
is positive on both receiver types in five. An unresolved router $D$
therefore points to the interface itself, not an insensitive contrast. Qwen3.6
is the exception: its residual $D$ is positive on error receivers
($0.04\;[0.01,0.06]$) but not on correct receivers ($-0.01\;[-0.03,0.01]$),
a direct difference of $-0.05\;[-0.07,-0.02]$, so its $D_{\mathrm{common}}$ of
$0.01\;[-0.01,0.03]$ shows no consistent direction
(Appendix~\ref{app:local-d}).

Correct and erroneous donors also differ in content. A same-label placebo
compares two donors with the same correctness label, ordered by a TF-IDF text
classifier's error score, and asks whether this surface difference
reproduces residual $D$ (Appendix~\ref{app:surface-routing}). Six of eight
eligible checkpoints pass all three pre-specified placebo criteria. Qwen3.6
lacks a resolved residual $D$, and in OLMoE the placebo reaches about
two-thirds of real $D$, so a substantial part of OLMoE's effect is associated
with surface content. The supported reading is dependence on donor
correctness beyond this lexical proxy, not pure correctness. Local residual
$D$ and all-layer router $D$ also differ in layer and token coverage, so
their contrast is not a matched-coverage interface comparison.

\subsection{Natural errors on ProcessBench}
\label{sec:processbench}

The single-layer interface ordering also holds for naturally occurring
errors. In an exploratory ProcessBench analysis of GPT-OSS-20B, Gemma, and
Nemotron-3 Nano on 64 cross-problem pairs, residual replacements produce
larger absolute changes than router-logit replacements in all six
checkpoint-by-direction comparisons, and the residual-minus-router $T$
difference is positive in all three checkpoints. This analysis includes
neither $D$ nor token windows (Appendix~\ref{app:processbench-details}).

\section{Related Work}

\textbf{Error signals and step-level verification.}
Process supervision and fine-grained reasoning benchmarks motivate step-level
verification
\citep{lightman2024letsverify,song2025prmbench,zheng2025processbench}.
Prior work shows that internal representations encode information about
truthfulness, reasoning correctness, or answerability that is not always
reflected in model outputs
\citep{orgad2025knowmore,yuan2026hiddenerror,du2026recognitionrefusal}, and
intermediate-state probes have been used for self-verification and early exit
\citep{zhang2025selfverification}, though \citet{yuan2026hiddenerror} report
limited gains when steering final-answer accuracy. Other work intervenes on
verifier representations directly, such as steering verifier strictness
\citep{zhou2026verifysteer}. Bridging these threads, we evaluate fixed step
prefixes under a shared verification margin, testing not only whether step
correctness is readable in MoE routing, but also whether replacing router,
MoE-output, or residual activations shifts the judgment and whether that shift
depends on whether the donor step is correct.

\textbf{MoE routing and intervention.}
MoE routing reflects semantic structure \citep{olson2025semanticrouting} and
hidden-state geometry \citep{wang2026mythspecialization}, while other studies
analyze expert specialization, the upstream formation of routing decisions,
and experts that are important for generation
\citep{herbst2026expertstrikes,li2026crosslayer,su2026superexperts}.
Intervening on experts or routing can also change model behavior
\citep{fayyaz2026steermoe,bandarkar2026multilingual}, and recent work has begun
to separate routing diagnostics from behavioral influence under ablation,
quantization, or controlled gate changes
\citep{engmann2026causalaudit,gu2026routeflip,caruzzo2026routingdivergence}.
Related analyses study router-visible subspaces, expert allocation, factual
recall, and expert selectivity
\citep{ye2026polysemantic,yoon2026misrouted,lu2026expertaware,salomone2026modularity}.
Whereas these studies establish what routing encodes and that editing it alters
behavior, we test whether a routing edit shifts a step-level verification
judgment by transferring the correctness signal that routing encodes, comparing
router-logit, MoE-output, and post-MoE residual interventions across layer
coverage and token position under a shared verification margin.

\textbf{Interpreting activation replacement.}
Causal tracing and Patchscopes study model behavior through activation or
representation replacement
\citep{vig2020causalmediation,meng2022rome,ghandeharioun2024patchscopes}.
A central limitation of such analyses is that information readable from a
representation need not be causally used
\citep{hewitt2019controltasks,elazar2021amnesic,sharma2026decodability}, while
patching conclusions can depend on the intervention construction, evaluation
metric, and the layers and token positions patched
\citep{zhang2024activationpatching}. Subspace interventions and causal
localization introduce further interpretational pitfalls
\citep{makelov2024subspaceillusion,hase2023localization}. To address these
concerns directly, we pair decodability probes with two separate intervention
measures, the paired shift size ($T$) and the fixed-receiver donor contrast
($D$), evaluated across interfaces, layers, and token windows without
interpreting either quantity as a pathway contribution or causal fraction.\looseness=-1

\section{Limitations}
\label{sec:limitations}

As with any activation-replacement study, swapping internal states can move
activations off their natural distribution, and effect sizes depend on the
intervention site, scoring metric, and amount of computation replaced
\citep{zhang2024activationpatching,makelov2024subspaceillusion}. Our protocol
mitigates these concerns by pairing norm-matched random controls with the
fixed-receiver contrast $D$ and checking sensitivity under total-logprob
scoring, though a positive $D$ can still reflect features correlated with step
correctness rather than correctness alone. Our same-label lexical placebo
confirms correctness dependence beyond TF-IDF surface cues in six of eight
eligible checkpoints (Appendix~\ref{app:surface-routing}), leaving finer
semantic and structural controls for future work.

Our interventions operate at the interface level across nine diverse
open-weight MoE checkpoints, establishing where correctness signals are
carried without yet localizing effects to individual experts or formal
mediation pathways. Extending \method{} to expert-level localization,
multi-step generation beyond fixed step prefixes, and additional MoE families
and verification benchmarks are natural directions for future work
(Appendix~\ref{app:claim-boundaries}).

\section{Conclusion}

We introduced \method{} to test whether MoE models use routing to carry
step-level correctness information into their verification judgments. Across
nine checkpoints, four linked results show that router logits encode step
correctness and all-layer routing swaps are causally potent, yet the resulting
shift rarely carries correctness. First, router logits are decodable for
correctness, with linear-probe AUROC from 0.80 to 0.88 in all eight checkpoints
with recorded activations and consistent gains over a text-only baseline.
Second, routing is not causally inert: although routing is the weakest
interface at a single layer, replacing router logits across all routed layers
shifts the verification margin in all nine checkpoints beyond a random
perturbation of the same norm, exceeding single-layer residual replacement in
GPT-OSS-120B. Token position also matters consistently, as moving an
eight-token window from the end of the step to the first divergence narrows the
residual-versus-routing gap in all nine checkpoints.

Third, this shift rarely tracks donor correctness. Holding the receiver fixed,
correct and erroneous donors produce indistinguishable router margins at a
single layer in all nine checkpoints, and under all-layer routing only
GPT-OSS-20B shows a resolved difference, with the 95\% upper bound of
$D_{\mathrm{common}}$ below 15\% of the all-layer routing shift in six
checkpoints (even though GPT-OSS-120B has the larger routing shift). Fourth,
this near-null router result is not due to an insensitive test: under the same
fixed-receiver comparison, even single-layer replacements yield positive $D$ in
eight of nine checkpoints for post-MoE residual states and five of nine for MoE
outputs, although part of OLMoE's residual effect reflects surface content.

Decodability and causal potency therefore do not imply correctness transfer: an
intervention that shifts a verification judgment need not do so by transferring
what the intervened state encodes. Evaluating routing, or any intervention
site, requires measuring correctness dependence on a fixed receiver alongside
effect size, and across layer and token coverage rather than at a single site.

\clearpage
\subsection*{Reproducibility statement}

Experiments record frozen dataset and model revisions, pair hashes, resolved
configurations, seeds, commits, validation-access flags, and immutable predictions.
Appendix~\ref{app:records} summarizes the archived campaign records and
checksums; the method and
control appendices specify estimands, paired resampling, identity checks, and
protocol controls. Appendix~\ref{app:wide-d} separates the primary full-step
all-router $D$ analysis, its total-logprob sensitivity, and post hoc direct
contrasts, including
their input-identity and recovery records. Code, configurations, pair manifests,
and non-weight artifacts are publicly available at
\url{https://github.com/yucheng-du/RouteVerify}. Recorded revisions
identify separately downloadable model weights.

\subsection*{AI use statement}

Generative AI tools, including OpenAI Codex, Claude Code, and Cursor, assisted
with research framing and experimental design; implementation, testing, data selection and
reformatting, analysis, and interpretation; and literature review, figure
preparation, translation, and manuscript drafting and editing. They were not used
to generate study datasets or proofs, and external audit outputs were treated as
issue leads rather than evidence. AI-assisted outputs were checked against source
code, tests, machine-readable artifacts, and cited sources; no AI system had
authority to accept results or determine the scientific conclusions. The authors
take responsibility for the final text, claims, code, and artifacts.\looseness=-1

\bibliography{references}
\bibliographystyle{iclr2027_conference}

\appendix
\makeatletter
\setlength{\@fptop}{0pt}
\setlength{\@fpsep}{8pt plus 1pt minus 1pt}
\setlength{\@fpbot}{0pt plus 1fil}
\makeatother
\setcounter{topnumber}{4}
\setcounter{bottomnumber}{4}
\setcounter{totalnumber}{8}
\renewcommand{\topfraction}{0.90}
\renewcommand{\bottomfraction}{0.80}
\renewcommand{\textfraction}{0.05}
\renewcommand{\floatpagefraction}{0.75}
\section{Model and Verification Setup}
\label{app:model-protocols}

\subsection{Prompt and verification endpoint}

\looseness=-1
The verification prompt contains the mathematical problem, the solution trace
through the evaluated step, and a fixed request to choose between
\texttt{CORRECT} and \texttt{INCORRECT}. Dataset labels, pair identities,
generator metadata, and later solution steps are not shown to the model. The
scorer computes the mean conditional log probability of each complete candidate
string using the model's native tokenizer and chat template. Intervention runs
retain the receiver prompt and modify only the designated internal activation.

\looseness=-1
PRMBench pairs share a problem and an identical prefix before the first
benchmark-annotated error step. The nine-checkpoint intervention analyses use
290 ordered pairs grouped into 235 canonical source-problem components;
ProcessBench is analyzed separately with 64 cross-problem pairs. Each archived
analysis records its sample count, pair identity, bootstrap identity, and
source-prediction path.

We use the frozen \texttt{hitsmy/PRMBench\_Preview} snapshot recorded in the
release manifests. Eligible pairs have matching nonempty problems, equal
nonzero step counts, valid error indices with an unambiguous indexing base,
identical text before the first annotated error, and a changed step at that
boundary. Only complete, unique pairs are retained. An error-type-stratified 50/50
split of 580 complete pair groups (seed 20260730) assigns 290 to discovery
and 290 to validation, and the nine-checkpoint intervention experiments use
only the discovery partition. The 235 source components are bootstrap units,
not the units of this split. Labels follow the benchmark annotations without
independent mathematical review of every trace and error boundary.

\paragraph{Category-exclusion sensitivity.}
In a post hoc analysis, we exclude the six circular and three redundancy
pairs in the Simplicity category, retaining the original predictions,
donor assignments, and resampling protocol (seed 20260908). Full-step and
last-token panels retain 281 pairs in 230 components; eight-token contrasts
use their shared eligible subsets. All 72 single-layer residual-minus-router
and residual-minus-MoE-output $T$ intervals remain positive, and all nine
divergence-8-minus-suffix-8 gap intervals remain negative. For OLMoE, the
full-step all-router paired-minus-norm-random contrast changes from
$0.0466\;[0.0003,0.0938]$ to $0.0450\;[-0.0013,0.0928]$: the point estimate
barely moves, but the interval now includes zero. The full sample remains the
primary analysis.

\subsection{Models and intervention layers}

The nine evaluated checkpoints are Qwen3.6-35B-A3B, Qwen3-30B-A3B,
GLM-4.7-Flash, Nemotron-3.5 Lightning 30B-A3B-BF16, Nemotron-3 Nano
30B-A3B, Gemma 4 26B-A4B, OLMoE-1B-7B, and GPT-OSS-20B/120B, with the
following model identifiers.

\begingroup
\raggedright
\href{https://huggingface.co/google/gemma-4-26B-A4B-it}{\nolinkurl{google/gemma-4-26B-A4B-it}},
\href{https://huggingface.co/nvidia/NVIDIA-Nemotron-3-Nano-30B-A3B-BF16}{\nolinkurl{nvidia/NVIDIA-Nemotron-3-Nano-30B-A3B-BF16}},
\href{https://huggingface.co/nvidia/NVIDIA-Nemotron-3.5-Lightning-30B-A3B-BF16}{\nolinkurl{nvidia/NVIDIA-Nemotron-3.5-Lightning-30B-A3B-BF16}},
\href{https://huggingface.co/Qwen/Qwen3.6-35B-A3B}{\nolinkurl{Qwen/Qwen3.6-35B-A3B}},
\path{Qwen/Qwen3-30B-A3B},
\path{zai-org/GLM-4.7-Flash},
\path{allenai/OLMoE-1B-7B-0924-Instruct},
\path{openai/gpt-oss-20b}, and
\path{openai/gpt-oss-120b}.
Exact model and tokenizer revisions are retained in the resolved configurations
and artifact index.
\par
\endgroup

\begin{table*}[tbp]
\centering
\caption{\looseness=-1 Layer coverage in the nine-checkpoint analysis. Layers are zero-indexed
absolute decoder-block indices. ``All routed sites'' counts routed blocks
replaced by an all-router or all-layer MoE-output intervention, not every
decoder block. ``Probe layer'' refers only to completed representation analyses
in Appendix~\ref{app:access}.}
\label{tab:abc-anchors}
\tablestyle
\begin{tabular}{@{}lrrrr@{}}
\toprule
Model & Local $D$ layer & Window layer & All routed sites & Probe layer \\
\midrule
Qwen3.6-35B-A3B & 20 & 20 & 40 & 20 \\
Qwen3-30B-A3B & 36 & 24 & 48 & 24 \\
GLM-4.7-Flash & 24 & 24 & 46 & 24 \\
Nemotron-3.5 Lightning & 27 & 27 & 23 & 27 \\
Nemotron-3 Nano & 27 & 27 & 23 & 27 \\
Gemma 4 26B-A4B & 15 & 15 & 30 & 15 \\
OLMoE-1B-7B & 8 & 8 & 16 & --- \\
GPT-OSS 20B & 12 & 12 & 24 & 12 \\
GPT-OSS 120B & 18 & 18 & 36 & 18 \\
\bottomrule
\end{tabular}
\end{table*}

\subsection{Architecture-specific intervention boundaries}
\label{app:architecture-boundaries}

Router-logit interventions replace the adapter's pre-dispatch logits while
retaining native expert selection and weighting. MoE-output interventions
replace the routed contribution at the architecture-specific output boundary,
and post-MoE residual interventions replace the corresponding residual
activation.

For Qwen3.6 and GLM, the MoE-output boundary precedes addition of the shared
branch, so the shared-expert contribution remains receiver-derived. GLM has
47 decoder blocks, with dense block 0 excluded from the 46 routed sites. In
both Nemotron checkpoints, the L27 output boundary is the final routed-only
aggregate at \texttt{fc2\_latent\_proj}; the receiver shared-expert branch
remains unchanged, and the 23 routed expert blocks lie within 52 hybrid decoder
blocks.

\looseness=-1
Gemma replaces the normalized routed contribution at
\texttt{post\_feedforward\_layernorm\_2} while preserving its parallel dense
branch, and its residual boundary is post-FFN. OLMoE exposes the routed
aggregate before residual addition. Qwen3 and the GPT-OSS checkpoints expose
the native MoE/MLP-module output; GPT-OSS uses 24 and 36 routed blocks for the
20B and 120B checkpoints, respectively, under native MXFP4 execution and
routing rules. The executed adapters and resolved configurations define the
exact intervention boundaries.

\section{Intervention Protocol and Controls}
\label{app:nine-model-protocols}

\subsection{Fixed-receiver correctness contrast and donor matching}

For receiver pair $i$, we select an external matched donor pair $d(i)$ from a
different canonical source component. Matching prioritizes agreement in error
type, followed by proximity of the first-error position, with a seeded pair
hash used only for tie-breaking. The same external pair supplies its correct
and error activations to both receivers. Donor assignments remain fixed during
analysis; the 290 receiver pairs use 204 distinct external donor pairs.

Let $M_i^{r\leftarrow d,s}$ denote the verification margin after replacing the
designated last-token receiver activation with the corresponding activation
from donor role $s$, for receiver role $r\in\{\clean,\error\}$. The four
donor/receiver combinations give
\begin{align}
D_{i,r} &= M_i^{r\leftarrow d,\error}
          -M_i^{r\leftarrow d,\clean}, \\
D_{i,\mathrm{common}} &= \tfrac12(D_{i,\clean}+D_{i,\error}).
\label{eq:appendix-d-common}
\end{align}

\looseness=-1
Because the receiver is identical in the two terms defining $D_{i,r}$, its
baseline verification margin cancels. Receiver roles are averaged within pair
before computing $D_{\mathrm{common}}$, so the sample size remains 290 rather
than 580, and direct interface contrasts are likewise computed within pair
before resampling. A positive $D$ rules out a donor-role-independent
perturbation as a sufficient explanation for the comparison, but donor labels
remain correlated with text, content, and activation geometry. This local
experiment uses one layer and the final step token; the full-step all-router
analysis is reported in Appendix~\ref{app:wide-d}.

\subsection{Token windows and alignment}

The window experiment crosses four intervention conditions (single-layer
router logits, single-layer MoE output, single-layer residual, and all-layer
router logits) with four token regions: the final token of the evaluated step,
the final $k$ tokens, $k$ tokens centered on the first token-level divergence
between the paired rendered prefixes, and the full evaluated step.

For each tokenizer, we choose the largest $k\in\{4,8,16,32\}$ for which both
paired steps contain at least $k$ tokens in at least 90\% of discovery pairs,
using token lengths only. All nine checkpoints select $k=8$. Divergence is
defined as the first difference in the complete rendered causal prefixes. The
centered window is clipped to remain within the evaluated step without
shortening it, and an example whose first divergence lies outside that step
is ineligible for this condition. For even $k$, the divergence token occupies
the left of the two central positions before clipping. Suffix and divergence
windows may overlap or coincide.

Full-step replacement covers every receiver-step token even when the donor step
has a different length. If receiver and donor windows have lengths $R$ and $L$,
receiver offset $j$ maps to donor offset
\[
\left\lfloor j(L-1)/(R-1)+1/2\right\rfloor;
\]
when $R=1$, the final donor token is used. The mapping preserves endpoint
ordering and may reuse donor positions. It is an operational alignment rather
than a claim of semantic token equivalence.

\subsection{Control conditions}
\label{app:controls-full}

Each interface/window combination includes paired-donor, self-patch,
matched-external-donor, and norm-random conditions. Self and matched conditions
inherit the paired receiver positions; a matched donor supplies its own
corresponding window, and unavailable donor windows are recorded as missing.

\paragraph{Identity controls.}
Unpatched reruns and self-patches, which replace receiver activations with
their own recorded values, verify that the intervention machinery preserves
the receiver when no cross-example replacement is made. Failure of an identity
condition would invalidate the corresponding cross-example rows. Passing the
check validates the implementation; it is not evidence that the natural causal
contribution of the intervention site is zero.

\paragraph{Matched external donors.}
Matched donors follow the matching rule above and supply the same donor roles
as the corresponding paired-donor condition. They test whether the paired
donor produces a larger response than a donor from another problem; the
primary comparison is the direct paired-minus-matched difference. A relative
comparison, when reported in the archived artifacts, divides the difference of
means by the paired-donor effect; it is not a mean of pairwise ratios or the
fraction of an underlying correctness mechanism. Matched donors do not remove
the gap-compression ambiguity of Equation~\ref{eq:gap-compression}, because
donor and receiver roles still change together; the fixed-receiver contrast
addresses that question separately, and neither analysis decomposes $T$ into
mechanistic components.

\paragraph{Norm-random controls.}
\looseness=-1
The norm-random intervention adds a Gaussian direction to the receiver
activation, scaled to the $\ell_2$ norm of the paired donor--receiver
displacement separately at each intervention position and layer. It tests
whether a perturbation of similar magnitude but random direction produces a
comparable response. A smaller norm-random effect does not exclude a
structured but correctness-independent replacement, and equal token count
does not imply equal activation displacement or equal cumulative layer
coverage.

\paragraph{Native routing execution.}
The intervention adapters retain each checkpoint's native expert selection,
normalization, and dispatch after router-logit replacement. Execution
artifacts record patched logits together with the downstream expert IDs and
weights used for dispatch, and machine checks compare these quantities under
each checkpoint's native top-$k$ and normalization rules. These checks
validate the manipulation; they are not cross-checkpoint effect-size
covariates.

\paragraph{Eligibility.}
A pair contributes to $T$ only when both replacement directions are available.
Direct control and cross-window contrasts use the intersection of their eligible
pair sets, and unavailable windows are never treated as zero.
Table~\ref{tab:abc-window-n} reports the resulting sample sizes; last-token and
full-step panels use all 290 pairs and 235 source components.

\begin{table*}[tbp]
\centering
\caption{Eligible pairs/source components ($N/K$) for the eight-token windows.
Paired-donor, self-patch, and norm-random conditions share the paired mask.
``Shared'' is the suffix/divergence intersection used for the equal-length
position contrast. Matched-donor comparisons use their own complete-pair
intersection.}
\label{tab:abc-window-n}
\tablestyle
\begin{tabular}{@{}lccccc@{}}
\toprule
& \multicolumn{2}{c}{Suffix-8} & \multicolumn{2}{c}{Divergence-8} & Shared \\
\cmidrule(lr){2-3}\cmidrule(lr){4-5}
Model & Paired & Matched & Paired & Matched & Paired \\
\midrule
Qwen3.6 & 278/226 & 265/217 & 276/224 & 261/213 & 276/224 \\
Qwen3 & 278/226 & 265/217 & 276/224 & 261/213 & 276/224 \\
GLM-4.7-Flash & 274/222 & 256/209 & 272/220 & 252/205 & 272/220 \\
Lightning & 278/226 & 265/217 & 276/224 & 261/213 & 276/224 \\
Nano & 278/226 & 265/217 & 276/224 & 261/213 & 276/224 \\
Gemma 4 & 279/226 & 268/220 & 277/224 & 264/216 & 277/224 \\
OLMoE & 271/220 & 250/205 & 271/220 & 250/205 & 271/220 \\
GPT-OSS 20B & 272/220 & 253/206 & 271/219 & 251/204 & 271/219 \\
GPT-OSS 120B & 272/220 & 253/206 & 271/219 & 251/204 & 271/219 \\
\bottomrule
\end{tabular}
\end{table*}

\subsection{All-layer MoE-output intervention}

The all-layer MoE-output experiment replaces the last-token routed output at
every routed site listed in Table~\ref{tab:abc-anchors}. It includes paired
donors, matched external donors, and norm-random controls, with the
norm-random displacement matched independently at each routed site.
Paired-minus-matched and paired-minus-random contrasts are computed within pair
before aggregation. This intervention differs in both interface and coverage
from a single-layer MoE-output intervention and a full-step all-router
intervention, and it has no fixed-receiver $D$ experiment.

\subsection{Statistical analysis}
\label{app:statistics}

The nine-checkpoint window analysis reconstructs 161,480 pair-level $T$ values
from the saved scores and covers all four token windows, intervention
interfaces, controls, direct interface comparisons, cross-window contrasts, and
gap-change contrasts. Each of the 50,000 PCG64 bootstrap draws samples eligible
source components with replacement, retains all pairs belonging to each sampled
component, and computes a pair-weighted mean. Analyses sharing an eligibility
mask use the same resamples, and direct differences are formed within pair
before resampling.

Intervals are two-sided pointwise 95\% percentile intervals. Inference
conditions on the fixed checkpoints, discovery pool, and donor assignments;
reused donors are not independently resampled. No multiplicity correction or
population-of-models test is applied. Exact seeds and analysis identifiers are
stored with the archived analysis artifacts.

\section{Representation-Probe Analysis}
\label{app:access}

\looseness=-1
Router logits separate correct from erroneous steps in all eight probed
checkpoints, slightly less well than residual states and router inputs and
better than expert-aligned routing weights (Table~\ref{tab:interface-access}).
This readability does not predict intervention effects. GLM, for example, has a
router-logit AUROC of 0.818 but an unresolved local router $D$ and a positive
full-step all-router $T$, and GPT-OSS-120B carries a strong correct/error
signal in its router logits while its broad routing intervention behaves
differently from its local residual intervention.

\subsection{Probe protocols and checkpoint details}

\looseness=-1
The probes are L2-regularized logistic classifiers with training-fold
standardization and out-of-fold predictions. Folds group the 290 paired examples
by canonical source-problem component, so that linked examples are not split
across training and evaluation folds; for the six original checkpoints this
mapping was made post hoc, so that panel is retrospective. The positive class is
the erroneous step. AUROC measures how well a representation separates
correct from erroneous steps; it does not measure causal use of that
information or identify a correctness-specific representation. We probe
residual states, router inputs, router logits, and expert-aligned routing
representations.

\begin{table*}[tbp]
\centering
\renewcommand{\arraystretch}{1.10}
\caption{Middle-anchor matched out-of-fold AUROC under 235 canonical
source-problem groups. ``Access/causal layer'' gives the observational recapture
anchor and the headline intervention anchor, respectively. ``Expert weight'' is
the expert-ID-aligned sparse gate-weight representation. GLM and GPT-OSS-120B
were probed later with the same folds; their intervals and direct contrasts are
given in the text. These are access measurements at capture interfaces, not
intervention effects or a mediation sequence.}
\label{tab:interface-access}
\tablestyle
\begin{tabular}{@{}llccccc@{}}
\toprule
Model & \shortstack{Protocol\\grade} & \shortstack{Access /\\causal layer} & Residual & \shortstack{Router\\input} &
\shortstack{Router\\logits} & \shortstack{Expert\\weight} \\
\midrule
Gemma 4 & modern & 15 / 15 & 0.893 & 0.901 & 0.868 & 0.792 \\
Nemotron-3 Nano & modern & 27 / 27 & 0.855 & 0.865 & 0.831 & 0.688 \\
Nemotron-3.5 Lightning & modern & 27 / 27 & 0.865 & 0.862 & 0.826 & 0.733 \\
Qwen3.6-35B-A3B & modern & 20 / 20 & 0.871 & 0.873 & 0.815 & 0.790 \\
GPT-OSS 20B & extended & 12 / 12 & 0.829 & 0.837 & 0.801 & 0.715 \\
Qwen3-30B-A3B & historical retrospective & 24 / 36 & 0.907 & 0.915 & 0.884 & 0.750 \\
OLMoE-1B-7B & unavailable & --- / 8 & --- & --- & --- & --- \\
\midrule
GLM-4.7-Flash & added later & 24 / 24 & 0.865 & 0.872 & 0.818 & 0.743 \\
GPT-OSS 120B & added later & 18 / 18 & 0.881 & 0.885 & 0.860 & 0.704 \\
\bottomrule
\end{tabular}
\end{table*}

Across the six original checkpoints, residual and router-input representations
generally have higher AUROC than router logits, which in turn exceed
expert-aligned sparse-weight representations. A narrower 238-group
recapture-native construction changes every cell by at most 0.031 and preserves
this ordering. Qwen3 uses a different probe/intervention layer pair (L24/L36) and
is therefore not an equal-layer intervention comparison. OLMoE has no
corresponding recapture manifest or activation tensors and is omitted rather
than assigned an imputed value. No additional contrast-feature control is
constructed, because the archived manifests contain no predeclared topic,
language, token-class, or formatting labels.

\paragraph{GLM and GPT-OSS-120B.}
These two checkpoints were probed later at L24 and L18 with the same 290 pairs,
580 rows, 235 source components, and five grouped folds. Their router-logit
AUROCs are $0.818\;[0.786,0.848]$ and $0.860\;[0.831,0.887]$, and their
expert-membership AUROCs $0.701$ and $0.670$. Router inputs exceed router
logits in both ($0.054\;[0.029,0.078]$ and $0.025\;[0.005,0.045]$), while
residual-minus-input differences are unresolved ($-0.007\;[-0.016,0.002]$ and
$-0.003\;[-0.011,0.004]$). In GPT-OSS-120B, router logits also exceed aligned
weights ($0.156\;[0.113,0.192]$), which exceed expert membership
($0.034\;[0.021,0.047]$). These probe intervals condition on the fitted
out-of-fold predictions and use 1,000 source-component draws with seed
20260829, separately from the intervention analyses. GLM's separate 64-pair
behavior subset has AUROC $0.840$.

The GPT-OSS-20B and GPT-OSS-120B recaptures share raw prompt content, pair
identities, and verification boundaries but differ in the rendered Harmony
system-header date, so their rendered-prompt and token-input hashes differ for
all 580 rows. Together with checkpoint and architecture differences, this
prevents interpreting their probe difference as an effect of model scale.

\paragraph{Text baseline on the same folds.}
A fixed word unigram/bigram TF--IDF baseline uses the boundary-step text and the
same L2-logistic settings as the router probes ($C=1$, balanced classes,
\texttt{liblinear}, 2,000 maximum iterations), without tuning. Vocabulary and
IDF are fit only within each training fold, and the comparison uses all 580
samples and the same five folds. Because the text inputs are identical across
the five paired-first-error protocols, one fitted baseline is reused across
checkpoints. Table~\ref{tab:text-baseline} compares router logits with this
baseline; intervals jointly resample the 235 source components 1,000 times
(seed 20260829) without refitting the classifiers, and Qwen3's distinct
boundary construction is kept separate.

\begin{table}[tbp]
\centering
\caption{Boundary-step text baseline on the identical 580 samples, 235 source components, and five saved interface-access folds. Differences compare the middle router-logit probe with the fixed text baseline; 95\% intervals use paired source-component resampling.}
\label{tab:text-baseline}
\tablestyle
\begin{tabular}{@{}lcc@{}}
\toprule
Model (access layer) & Text AUROC [95\% CI] & Router minus text [95\% CI] \\
\midrule
Gemma 4 (L15) & \(0.743\)\hspace{0.25em}\ci{\([0.710, 0.777]\)} & \({+}0.124\)\hspace{0.25em}\ci{\([0.084, 0.159]\)} \\
Nemotron-3 Nano (L27) & \(0.743\)\hspace{0.25em}\ci{\([0.710, 0.777]\)} & \({+}0.088\)\hspace{0.25em}\ci{\([0.051, 0.123]\)} \\
Nemotron-3.5 Lightning (L27) & \(0.743\)\hspace{0.25em}\ci{\([0.710, 0.777]\)} & \({+}0.083\)\hspace{0.25em}\ci{\([0.043, 0.116]\)} \\
Qwen3.6-35B-A3B (L20) & \(0.743\)\hspace{0.25em}\ci{\([0.710, 0.777]\)} & \({+}0.072\)\hspace{0.25em}\ci{\([0.031, 0.110]\)} \\
GPT-OSS-20B (L12) & \(0.743\)\hspace{0.25em}\ci{\([0.710, 0.777]\)} & \({+}0.057\)\hspace{0.25em}\ci{\([0.024, 0.092]\)} \\
\midrule
\multicolumn{3}{@{}l}{\emph{Historical boundary sensitivity}} \\
Qwen3-30B-A3B (L24) & \(0.734\)\hspace{0.25em}\ci{\([0.696, 0.771]\)} & \({+}0.150\)\hspace{0.25em}\ci{\([0.110, 0.190]\)} \\
\bottomrule
\end{tabular}
\end{table}

\begingroup\parfillskip=0pt plus .5\linewidth\finalhyphendemerits=100000
A router advantage over this baseline does not establish a correctness-specific
representation or remove all surface confounds: contextual router states contain
information unavailable to isolated boundary-step word features, and the
representation dimensions and classifier capacities are not matched. In
Qwen3's separate matched surface comparison, routing and text features reach
AUROCs of 0.777 and 0.795, with paired difference $-0.019\;[-0.047,0.010]$,
which is compatible with readable routing information but does not establish
that it is used.
\par\endgroup

The probes passively read representations, whereas the intervention analyses
replace activations at adapter-defined boundaries and measure changes in the
verification margin. Similar interface names do not make these measurements
stages of a serial mediation pathway.

\section{Supplementary Results for Judgment Changes}
\label{app:nine-model-boundaries}

\begin{figure}[tbp]
\centering
\includegraphics[width=\textwidth]{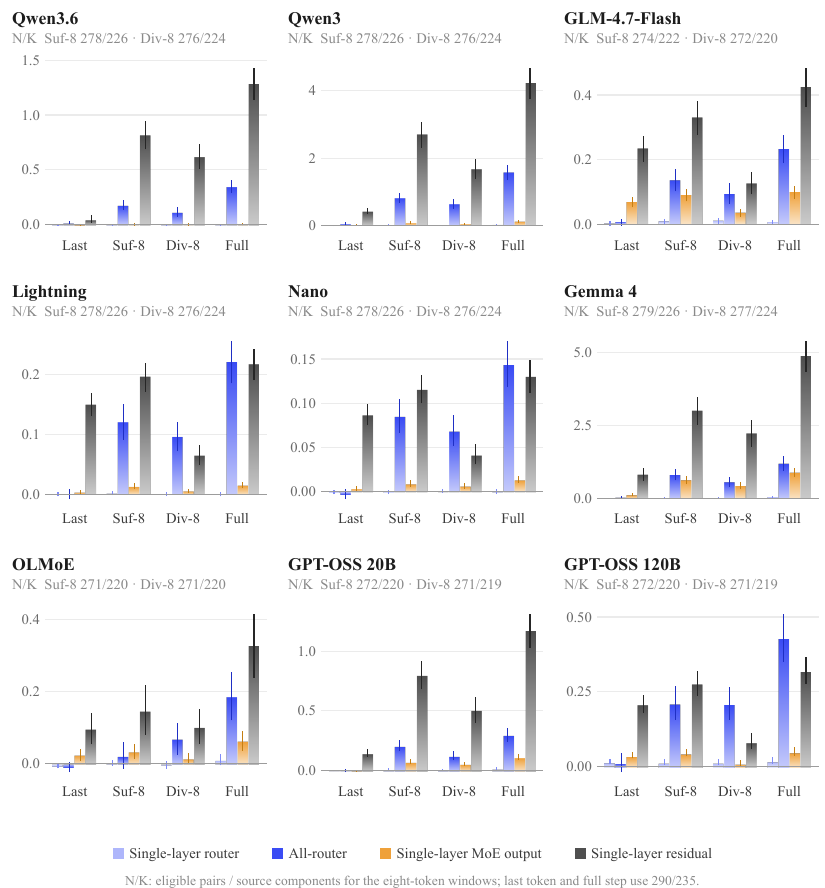}
\caption{Nine-checkpoint intervention profiles across four token windows and
four intervention conditions, on checkpoint-specific linear axes. Bars show
pair-weighted $T$ means and whiskers 95\% intervals; eight-token sample sizes
are given in Table~\ref{tab:abc-window-n}. Local intervention layers are fixed
within each checkpoint, whereas the all-router condition replaces router logits
at every routed layer.}
\label{fig:abc-window-profiles}
\end{figure}

This appendix reports the per-checkpoint $T$ profiles behind
Figure~\ref{fig:abc-windows} and the controls behind Section~\ref{sec:windows}
(Figure~\ref{fig:abc-window-profiles} and
Tables~\ref{tab:abc-window-router-controls}--\ref{tab:abc-bundle}).

\paragraph{Coverage and position.}
Expanding token coverage from the last token to the full step does not change
the residual-minus-all-router gap consistently: the gap increases in Qwen3.6,
Qwen3, Gemma, and GPT-OSS-20B; decreases in GLM, Lightning, Nano, and
GPT-OSS-120B; and is unresolved in OLMoE. Position is more consistent. With
token count fixed at eight, moving from suffix-8 to divergence-8 reduces the gap
in all nine checkpoints. This does not imply a larger absolute routing response
near divergence: in GPT-OSS-120B, the all-router divergence-minus-suffix
difference is $-0.003\;[-0.059,0.054]$, whereas the residual difference is
$-0.195\;[-0.232,-0.159]$, so the changed ordering there comes mainly from
residual sensitivity to token position.

\begingroup\parfillskip=0pt plus .5\linewidth\finalhyphendemerits=100000
Qwen3.6's full-step residual effect also differs from its heavy-tailed
last-token result: its mean is $1.28$, its median $1.21$, its 10\%-trimmed mean
$1.22$, and 88.3\% of pair effects are positive. The pair-weighted mean remains
the primary summary.
\par\endgroup

\begin{table*}[tbp]
\centering
\caption{\looseness=-1 Controls for full-step all-router interventions and directly paired
contrasts. Norm-random interventions match the paired donor displacement
separately at each token position and layer. OLMoE's ($\dagger$)
paired-minus-norm-random interval lies close to zero; a larger matched-donor
$T$ does not imply more correctness-related information.}
\label{tab:abc-window-router-controls}
\begingroup
\renewcommand{\arraystretch}{1.15}
\tablestyle
\begin{tabular}{@{}lccc@{}}
\toprule
Model & Paired \(T\) & Paired\(-\)matched & Paired\(-\)norm-random \\
\midrule
Qwen3.6 & \(\phantom{1}.347\)\hspace{0.25em}\ci{\([\phantom{1}.286, \phantom{1}.408]\)} & \({-}.168\)\hspace{0.25em}\ci{\([\phantom{{-}1}\llap{\({-}\)}.227, {-}.111]\)} & \(.218\)\hspace{0.25em}\ci{\([.170, \phantom{1}.267]\)}\\[1pt]
Qwen3 & \(1.576\)\hspace{0.25em}\ci{\([1.358, 1.796]\)} & \({-}.497\)\hspace{0.25em}\ci{\([\phantom{{-}1}\llap{\({-}\)}.663, {-}.335]\)} & \(.651\)\hspace{0.25em}\ci{\([.505, \phantom{1}.804]\)}\\[1pt]
GLM-4.7-Flash & \(\phantom{1}.231\)\hspace{0.25em}\ci{\([\phantom{1}.189, \phantom{1}.275]\)} & \({-}.071\)\hspace{0.25em}\ci{\([\phantom{{-}1}\llap{\({-}\)}.102, {-}.040]\)} & \(.165\)\hspace{0.25em}\ci{\([.131, \phantom{1}.200]\)}\\[1pt]
Lightning & \(\phantom{1}.221\)\hspace{0.25em}\ci{\([\phantom{1}.186, \phantom{1}.256]\)} & \({-}.073\)\hspace{0.25em}\ci{\([\phantom{{-}1}\llap{\({-}\)}.096, {-}.050]\)} & \(.175\)\hspace{0.25em}\ci{\([.144, \phantom{1}.208]\)}\\[1pt]
Nano & \(\phantom{1}.143\)\hspace{0.25em}\ci{\([\phantom{1}.118, \phantom{1}.170]\)} & \({-}.040\)\hspace{0.25em}\ci{\([\phantom{{-}1}\llap{\({-}\)}.056, {-}.024]\)} & \(.120\)\hspace{0.25em}\ci{\([.098, \phantom{1}.143]\)}\\[1pt]
Gemma 4 & \(1.208\)\hspace{0.25em}\ci{\([\phantom{1}.958, 1.459]\)} & \({-}.926\)\hspace{0.25em}\ci{\([{-}1.199, {-}.655]\)} & \(.915\)\hspace{0.25em}\ci{\([.689, 1.141]\)}\\[1pt]
OLMoE\(^{\dagger}\) & \(\phantom{1}.185\)\hspace{0.25em}\ci{\([\phantom{1}.120, \phantom{1}.253]\)} & \({-}.017\)\hspace{0.25em}\ci{\([\phantom{{-}1}\llap{\({-}\)}.065, \phantom{{-}}.030]\)} & \(.047\)\hspace{0.25em}\ci{\([.000, \phantom{1}.094]\)}\\[1pt]
GPT-OSS-20B & \(\phantom{1}.297\)\hspace{0.25em}\ci{\([\phantom{1}.242, \phantom{1}.353]\)} & \({-}.161\)\hspace{0.25em}\ci{\([\phantom{{-}1}\llap{\({-}\)}.214, {-}.109]\)} & \(.189\)\hspace{0.25em}\ci{\([.146, \phantom{1}.233]\)}\\[1pt]
GPT-OSS-120B & \(\phantom{1}.428\)\hspace{0.25em}\ci{\([\phantom{1}.350, \phantom{1}.509]\)} & \({-}.169\)\hspace{0.25em}\ci{\([\phantom{{-}1}\llap{\({-}\)}.229, {-}.111]\)} & \(.214\)\hspace{0.25em}\ci{\([.155, \phantom{1}.275]\)}\\[1pt]
\bottomrule
\end{tabular}
\endgroup

\end{table*}

\paragraph{Control comparisons.}
Full-step all-router paired-minus-matched intervals are negative in eight
checkpoints and include zero in OLMoE, whereas paired-minus-norm-random
intervals are positive in all nine, with OLMoE close to zero ($0.047$
$[0.0003,0.094]$; norm-random $T=0.139$, paired-donor $T=0.185$). For the
all-layer MoE-output intervention (Table~\ref{tab:abc-bundle}), every
paired-minus-norm-random interval is positive and every paired-minus-matched
interval includes zero, which does not establish equivalence between paired and
matched external donors. These controls show effects beyond random
displacement magnitude without resolving paired-example specificity.

\begin{table}[tbp]
\centering
\caption{
All-layer MoE-output interventions at the last token. Paired-donor $T$ is
positive and exceeds the norm-matched random control in all nine checkpoints.
``Paired'' uses the corresponding correct/error trace as the donor, whereas
``matched'' uses an external donor from another problem matched by error type
and boundary position. All paired-minus-matched intervals include zero.}
\begingroup
\renewcommand{\arraystretch}{1.15}
\tablestyle
\begin{tabular}{@{}lccc@{}}
\toprule
Model & Paired-donor \(T\) & Paired\(-\)matched & Paired\(-\)norm-random \\
\midrule
Qwen3.6 & \(\phantom{1}.099\)\hspace{0.25em}\ci{\([.064, \phantom{1}.138]\)} & \(.019\)\hspace{0.25em}\ci{\([{-}.007, .048]\)} & \(.088\)\hspace{0.25em}\ci{\([.051, \phantom{1}.130]\)}\\[1pt]
Qwen3 & \(\phantom{1}.933\)\hspace{0.25em}\ci{\([.785, 1.083]\)} & \(.084\)\hspace{0.25em}\ci{\([{-}.022, .190]\)} & \(.848\)\hspace{0.25em}\ci{\([.713, \phantom{1}.984]\)}\\[1pt]
GLM-4.7-Flash & \(\phantom{1}.162\)\hspace{0.25em}\ci{\([.127, \phantom{1}.198]\)} & \(.014\)\hspace{0.25em}\ci{\([{-}.012, .040]\)} & \(.149\)\hspace{0.25em}\ci{\([.116, \phantom{1}.183]\)}\\[1pt]
Lightning & \(\phantom{1}.214\)\hspace{0.25em}\ci{\([.181, \phantom{1}.249]\)} & \(.019\)\hspace{0.25em}\ci{\([{-}.010, .048]\)} & \(.196\)\hspace{0.25em}\ci{\([.164, \phantom{1}.229]\)}\\[1pt]
Nano & \(\phantom{1}.101\)\hspace{0.25em}\ci{\([.083, \phantom{1}.120]\)} & \(.008\)\hspace{0.25em}\ci{\([{-}.006, .022]\)} & \(.088\)\hspace{0.25em}\ci{\([.071, \phantom{1}.106]\)}\\[1pt]
Gemma 4 & \(1.133\)\hspace{0.25em}\ci{\([.869, 1.422]\)} & \(.136\)\hspace{0.25em}\ci{\([{-}.107, .388]\)} & \(.938\)\hspace{0.25em}\ci{\([.684, 1.217]\)}\\[1pt]
OLMoE & \(\phantom{1}.113\)\hspace{0.25em}\ci{\([.067, \phantom{1}.162]\)} & \(.007\)\hspace{0.25em}\ci{\([{-}.035, .050]\)} & \(.093\)\hspace{0.25em}\ci{\([.055, \phantom{1}.135]\)}\\[1pt]
GPT-OSS-20B & \(\phantom{1}.388\)\hspace{0.25em}\ci{\([.331, \phantom{1}.446]\)} & \(.030\)\hspace{0.25em}\ci{\([{-}.015, .077]\)} & \(.364\)\hspace{0.25em}\ci{\([.280, \phantom{1}.447]\)}\\[1pt]
GPT-OSS-120B & \(\phantom{1}.563\)\hspace{0.25em}\ci{\([.478, \phantom{1}.650]\)} & \(.022\)\hspace{0.25em}\ci{\([{-}.045, .091]\)} & \(.528\)\hspace{0.25em}\ci{\([.446, \phantom{1}.612]\)}\\[1pt]
\bottomrule
\end{tabular}
\endgroup

\label{tab:abc-bundle}
\end{table}

\paragraph{Two GPT-OSS checkpoints.}
Matched-pair 120B-minus-20B contrasts are $0.13\;[0.07,0.19]$ for full-step
all-router $T$ and $-0.85\;[-0.97,-0.73]$ for full-step residual $T$.
Single-layer $D_{\mathrm{common}}$ differences are $0.045\;[0.029,0.062]$ at the
MoE-output interface and $0.077\;[0.041,0.114]$ at the residual interface.
Effects therefore do not grow uniformly across interfaces in the larger
checkpoint. The two checkpoints also differ in local anchor, routed depth,
expert count, training, and checkpoint identity, so these comparisons do not
isolate parameter count or define a scaling law.

\section{Verification Performance, Decision Flips, and Normalized Effects}
\label{app:verification-metrics}

All three quantities are computed from the saved candidate scores of the
nine-checkpoint window runs and, where stated, of the full-step all-router
fixed-receiver run, without additional model forwards. A receiver receives the
verdict \texttt{INCORRECT} when $\marginm>0$ and \texttt{CORRECT} otherwise; the
truth is \texttt{CORRECT} for the correct receiver and \texttt{INCORRECT} for
the error receiver. These are within-study values of a fixed candidate-scoring
verdict on our paired PRMBench assay, not official PRMBench scores. The
total-logprob margin gives the same picture as the primary mean-logprob margin
of Equation~\ref{eq:margin}, with accuracy within $0.021$ and AUROC within
$0.002$; exact ties at zero occur only on the total metric, in 5 baseline
receivers of Qwen3, and are read as \texttt{CORRECT}. AUROC intervals use
resample-weighted, tie-aware rank comparisons.

\subsection{Unpatched verification performance}

\looseness=-1
Accuracy is the fraction of the 580 receivers whose verdict matches the truth.
Error recall is the fraction of error receivers called \texttt{INCORRECT}, and
the correct-receiver false-positive rate is the fraction of correct receivers
called \texttt{INCORRECT}. AUROC ranks receivers by $\marginm$ with the error
receiver as the positive class, and the mean gap
$g=\marginm(x_\error)-\marginm(x_\clean)$ averages the within-pair margin
difference. Labels are balanced, so chance is $0.5$ for accuracy and AUROC, and
balanced accuracy equals accuracy.

All nine checkpoints are above chance (Table~\ref{tab:verification-baseline}),
with accuracy from $0.57$ in OLMoE to $0.79$ in Gemma and AUROC from $0.60$ to
$0.87$. Two checkpoints are strongly biased in opposite directions: Nano calls
$0.62$ of correct receivers \texttt{INCORRECT}, whereas OLMoE never does and
recalls only $0.13$ of errors. Unpatched reruns within the window runs reproduce
these scores exactly, with zero verdict flips in every checkpoint.

\begin{table*}[tbp]
\centering
\caption{Unpatched verification performance on our paired PRMBench assay,
under the mean-logprob margin and the $\marginm>0$ verdict rule.
Accuracy, error recall, and the correct-receiver false-positive rate use all 580
receivers of the 290 pairs; AUROC takes the error receiver as the positive
class; $g$ is the within-pair baseline margin gap. These are within-study assay
values, not official PRMBench benchmark scores.}
\label{tab:verification-baseline}
\begingroup
\renewcommand{\arraystretch}{1.15}
\tablestyle
\begin{tabular}{@{}lccccc@{}}
\toprule
Model & Accuracy & Error recall & \begin{tabular}[t]{@{}c@{}}Correct-receiver\\FPR\end{tabular} & AUROC & Mean gap \(g\) \\
\midrule
Qwen3.6 & \begin{tabular}[t]{@{}c@{}}\(0.7776\)\\\ci{\([0.7457, 0.8090]\)}\end{tabular} & \begin{tabular}[t]{@{}c@{}}\(0.6517\)\\\ci{\([0.5951, 0.7075]\)}\end{tabular} & \begin{tabular}[t]{@{}c@{}}\(0.0966\)\\\ci{\([0.0605, 0.1359]\)}\end{tabular} & \begin{tabular}[t]{@{}c@{}}\(0.8580\)\\\ci{\([0.8279, 0.8866]\)}\end{tabular} & \begin{tabular}[t]{@{}c@{}}\(1.5549\)\\\ci{\([1.3789, 1.7296]\)}\end{tabular}\\[2pt]
Qwen3 & \begin{tabular}[t]{@{}c@{}}\(0.7345\)\\\ci{\([0.7038, 0.7650]\)}\end{tabular} & \begin{tabular}[t]{@{}c@{}}\(0.5586\)\\\ci{\([0.5000, 0.6151]\)}\end{tabular} & \begin{tabular}[t]{@{}c@{}}\(0.0897\)\\\ci{\([0.0559, 0.1263]\)}\end{tabular} & \begin{tabular}[t]{@{}c@{}}\(0.8610\)\\\ci{\([0.8316, 0.8891]\)}\end{tabular} & \begin{tabular}[t]{@{}c@{}}\(4.6818\)\\\ci{\([4.1758, 5.1846]\)}\end{tabular}\\[2pt]
GLM-4.7-Flash & \begin{tabular}[t]{@{}c@{}}\(0.5845\)\\\ci{\([0.5621, 0.6075]\)}\end{tabular} & \begin{tabular}[t]{@{}c@{}}\(0.1862\)\\\ci{\([0.1424, 0.2319]\)}\end{tabular} & \begin{tabular}[t]{@{}c@{}}\(0.0172\)\\\ci{\([0.0035, 0.0339]\)}\end{tabular} & \begin{tabular}[t]{@{}c@{}}\(0.8058\)\\\ci{\([0.7748, 0.8360]\)}\end{tabular} & \begin{tabular}[t]{@{}c@{}}\(0.8147\)\\\ci{\([0.7125, 0.9175]\)}\end{tabular}\\[2pt]
Lightning & \begin{tabular}[t]{@{}c@{}}\(0.7621\)\\\ci{\([0.7298, 0.7937]\)}\end{tabular} & \begin{tabular}[t]{@{}c@{}}\(0.7517\)\\\ci{\([0.7013, 0.8007]\)}\end{tabular} & \begin{tabular}[t]{@{}c@{}}\(0.2276\)\\\ci{\([0.1771, 0.2800]\)}\end{tabular} & \begin{tabular}[t]{@{}c@{}}\(0.8475\)\\\ci{\([0.8185, 0.8755]\)}\end{tabular} & \begin{tabular}[t]{@{}c@{}}\(0.8680\)\\\ci{\([0.7716, 0.9655]\)}\end{tabular}\\[2pt]
Nano & \begin{tabular}[t]{@{}c@{}}\(0.6534\)\\\ci{\([0.6241, 0.6838]\)}\end{tabular} & \begin{tabular}[t]{@{}c@{}}\(0.9276\)\\\ci{\([0.8953, 0.9570]\)}\end{tabular} & \begin{tabular}[t]{@{}c@{}}\(0.6207\)\\\ci{\([0.5586, 0.6812]\)}\end{tabular} & \begin{tabular}[t]{@{}c@{}}\(0.8171\)\\\ci{\([0.7857, 0.8478]\)}\end{tabular} & \begin{tabular}[t]{@{}c@{}}\(0.5024\)\\\ci{\([0.4378, 0.5689]\)}\end{tabular}\\[2pt]
Gemma 4 & \begin{tabular}[t]{@{}c@{}}\(0.7862\)\\\ci{\([0.7552, 0.8165]\)}\end{tabular} & \begin{tabular}[t]{@{}c@{}}\(0.8103\)\\\ci{\([0.7651, 0.8537]\)}\end{tabular} & \begin{tabular}[t]{@{}c@{}}\(0.2379\)\\\ci{\([0.1852, 0.2941]\)}\end{tabular} & \begin{tabular}[t]{@{}c@{}}\(0.8744\)\\\ci{\([0.8470, 0.9003]\)}\end{tabular} & \begin{tabular}[t]{@{}c@{}}\(7.0863\)\\\ci{\([6.3870, 7.7836]\)}\end{tabular}\\[2pt]
OLMoE & \begin{tabular}[t]{@{}c@{}}\(0.5672\)\\\ci{\([0.5481, 0.5875]\)}\end{tabular} & \begin{tabular}[t]{@{}c@{}}\(0.1345\)\\\ci{\([0.0962, 0.1751]\)}\end{tabular} & \begin{tabular}[t]{@{}c@{}}\(0.0000\)\\\ci{\([0.0000, 0.0000]\)}\end{tabular} & \begin{tabular}[t]{@{}c@{}}\(0.6018\)\\\ci{\([0.5701, 0.6334]\)}\end{tabular} & \begin{tabular}[t]{@{}c@{}}\(0.6056\)\\\ci{\([0.4427, 0.7738]\)}\end{tabular}\\[2pt]
GPT-OSS-20B & \begin{tabular}[t]{@{}c@{}}\(0.7155\)\\\ci{\([0.6858, 0.7451]\)}\end{tabular} & \begin{tabular}[t]{@{}c@{}}\(0.8034\)\\\ci{\([0.7562, 0.8486]\)}\end{tabular} & \begin{tabular}[t]{@{}c@{}}\(0.3724\)\\\ci{\([0.3132, 0.4339]\)}\end{tabular} & \begin{tabular}[t]{@{}c@{}}\(0.8202\)\\\ci{\([0.7901, 0.8497]\)}\end{tabular} & \begin{tabular}[t]{@{}c@{}}\(1.2942\)\\\ci{\([1.1356, 1.4525]\)}\end{tabular}\\[2pt]
GPT-OSS-120B & \begin{tabular}[t]{@{}c@{}}\(0.7103\)\\\ci{\([0.6781, 0.7425]\)}\end{tabular} & \begin{tabular}[t]{@{}c@{}}\(0.8138\)\\\ci{\([0.7662, 0.8596]\)}\end{tabular} & \begin{tabular}[t]{@{}c@{}}\(0.3931\)\\\ci{\([0.3345, 0.4539]\)}\end{tabular} & \begin{tabular}[t]{@{}c@{}}\(0.7785\)\\\ci{\([0.7460, 0.8104]\)}\end{tabular} & \begin{tabular}[t]{@{}c@{}}\(1.3498\)\\\ci{\([1.1818, 1.5193]\)}\end{tabular}\\[2pt]
\bottomrule
\end{tabular}
\endgroup

\end{table*}

\subsection{Decision flip rates}

For each receiver we compare the unpatched with the patched verdict. The four
transition classes are correct-to-correct, correct-to-wrong, wrong-to-correct,
and wrong-to-wrong, where correct and wrong refer to agreement with the truth,
and the flip rate is the fraction of receivers whose verdict changes. The
harmful rate conditions on baseline-correct receivers and the beneficial rate
on baseline-wrong receivers; with different denominators, the two are never
subtracted, and the change in accuracy equals the unconditional beneficial rate
minus the unconditional harmful rate. Because the paired design supplies the
opposite-role donor, a donor-aligned flip is a flip away from the truth, and
accuracy decreases in every paired residual panel and every full-step
all-router panel. Where flips are few the sign is not forced: in 41 of the
remaining 198 paired panels, all router-logit, MoE-output, or non-full-step
all-router panels, accuracy instead rises by at most $0.0091$ (five receivers
or fewer). Positive mean $\teffect$ thus indicates an average margin shift
toward the donors' labels but does not determine the accuracy change.

\begingroup\parfillskip=0pt plus .5\linewidth\finalhyphendemerits=100000
At full step, the all-router flip rate is highest with matched donors and
lowest under the norm-random control in every checkpoint
(Table~\ref{tab:decision-flips}). Paired residual replacement separates the
checkpoints into two regimes. Qwen3.6, Qwen3, Gemma, and GPT-OSS-20B fall to
patched accuracy of $0.27$ to $0.37$, well below chance, whereas the other five
lose only $0.07$ to $0.10$ of accuracy and end near or above chance.
Single-layer router-logit replacement flips $0.005$ to $0.026$ of verdicts at
full step. Conditional beneficial rates are undefined for OLMoE on correct
receivers, whose false-positive rate is exactly zero, and rest on one to five
receivers for GLM-4.7-Flash; unconditional rates and accuracy changes are
defined throughout.
\par\endgroup

\begin{table*}[tbp]
\centering
\caption{Full-step verdict flips under the mean-logprob margin, on all 580
receivers of the 290 pairs. The first three columns give the all-router flip
rate under the paired donor, a matched external donor, and the norm-random
control; the last two give the paired residual flip rate and its accuracy
change. Self-patch controls produce zero flips in every checkpoint.}
\label{tab:decision-flips}
\begingroup
\renewcommand{\arraystretch}{1.15}
\tablestyle
\begin{tabular}{@{}lccccc@{}}
\toprule
Model & \begin{tabular}[t]{@{}c@{}}All-router\\paired\end{tabular} & \begin{tabular}[t]{@{}c@{}}All-router\\matched\end{tabular} & \begin{tabular}[t]{@{}c@{}}All-router\\norm-random\end{tabular} & \begin{tabular}[t]{@{}c@{}}Residual\\paired\end{tabular} & \begin{tabular}[t]{@{}c@{}}Residual\\\(\Delta\)accuracy\end{tabular} \\
\midrule
Qwen3.6 & \begin{tabular}[t]{@{}c@{}}\(0.1690\)\\\ci{\([0.1399, 0.1993]\)}\end{tabular} & \begin{tabular}[t]{@{}c@{}}\(0.2690\)\\\ci{\([0.2352, 0.3032]\)}\end{tabular} & \begin{tabular}[t]{@{}c@{}}\(0.0879\)\\\ci{\([0.0651, 0.1123]\)}\end{tabular} & \begin{tabular}[t]{@{}c@{}}\(0.5362\)\\\ci{\([0.4826, 0.5890]\)}\end{tabular} & \begin{tabular}[t]{@{}c@{}}\({-}0.5052\)\\\ci{\([{-}0.5629, {-}0.4467]\)}\end{tabular}\\[2pt]
Qwen3 & \begin{tabular}[t]{@{}c@{}}\(0.2155\)\\\ci{\([0.1831, 0.2491]\)}\end{tabular} & \begin{tabular}[t]{@{}c@{}}\(0.3517\)\\\ci{\([0.3134, 0.3903]\)}\end{tabular} & \begin{tabular}[t]{@{}c@{}}\(0.1121\)\\\ci{\([0.0850, 0.1406]\)}\end{tabular} & \begin{tabular}[t]{@{}c@{}}\(0.4397\)\\\ci{\([0.3867, 0.4929]\)}\end{tabular} & \begin{tabular}[t]{@{}c@{}}\({-}0.4155\)\\\ci{\([{-}0.4711, {-}0.3596]\)}\end{tabular}\\[2pt]
GLM-4.7-Flash & \begin{tabular}[t]{@{}c@{}}\(0.0966\)\\\ci{\([0.0722, 0.1224]\)}\end{tabular} & \begin{tabular}[t]{@{}c@{}}\(0.1155\)\\\ci{\([0.0896, 0.1421]\)}\end{tabular} & \begin{tabular}[t]{@{}c@{}}\(0.0586\)\\\ci{\([0.0402, 0.0782]\)}\end{tabular} & \begin{tabular}[t]{@{}c@{}}\(0.0983\)\\\ci{\([0.0732, 0.1242]\)}\end{tabular} & \begin{tabular}[t]{@{}c@{}}\({-}0.0879\)\\\ci{\([{-}0.1150, {-}0.0618]\)}\end{tabular}\\[2pt]
Lightning & \begin{tabular}[t]{@{}c@{}}\(0.2034\)\\\ci{\([0.1701, 0.2377]\)}\end{tabular} & \begin{tabular}[t]{@{}c@{}}\(0.3000\)\\\ci{\([0.2607, 0.3402]\)}\end{tabular} & \begin{tabular}[t]{@{}c@{}}\(0.0586\)\\\ci{\([0.0412, 0.0769]\)}\end{tabular} & \begin{tabular}[t]{@{}c@{}}\(0.1103\)\\\ci{\([0.0847, 0.1381]\)}\end{tabular} & \begin{tabular}[t]{@{}c@{}}\({-}0.0690\)\\\ci{\([{-}0.0956, {-}0.0423]\)}\end{tabular}\\[2pt]
Nano & \begin{tabular}[t]{@{}c@{}}\(0.1414\)\\\ci{\([0.1107, 0.1738]\)}\end{tabular} & \begin{tabular}[t]{@{}c@{}}\(0.1897\)\\\ci{\([0.1544, 0.2266]\)}\end{tabular} & \begin{tabular}[t]{@{}c@{}}\(0.0466\)\\\ci{\([0.0295, 0.0654]\)}\end{tabular} & \begin{tabular}[t]{@{}c@{}}\(0.1121\)\\\ci{\([0.0857, 0.1398]\)}\end{tabular} & \begin{tabular}[t]{@{}c@{}}\({-}0.1017\)\\\ci{\([{-}0.1301, {-}0.0742]\)}\end{tabular}\\[2pt]
Gemma 4 & \begin{tabular}[t]{@{}c@{}}\(0.1879\)\\\ci{\([0.1567, 0.2202]\)}\end{tabular} & \begin{tabular}[t]{@{}c@{}}\(0.3224\)\\\ci{\([0.2819, 0.3635]\)}\end{tabular} & \begin{tabular}[t]{@{}c@{}}\(0.0552\)\\\ci{\([0.0372, 0.0748]\)}\end{tabular} & \begin{tabular}[t]{@{}c@{}}\(0.4362\)\\\ci{\([0.3866, 0.4859]\)}\end{tabular} & \begin{tabular}[t]{@{}c@{}}\({-}0.4121\)\\\ci{\([{-}0.4653, {-}0.3582]\)}\end{tabular}\\[2pt]
OLMoE & \begin{tabular}[t]{@{}c@{}}\(0.0621\)\\\ci{\([0.0437, 0.0816]\)}\end{tabular} & \begin{tabular}[t]{@{}c@{}}\(0.0690\)\\\ci{\([0.0498, 0.0893]\)}\end{tabular} & \begin{tabular}[t]{@{}c@{}}\(0.0310\)\\\ci{\([0.0179, 0.0451]\)}\end{tabular} & \begin{tabular}[t]{@{}c@{}}\(0.0672\)\\\ci{\([0.0448, 0.0914]\)}\end{tabular} & \begin{tabular}[t]{@{}c@{}}\({-}0.0672\)\\\ci{\([{-}0.0914, {-}0.0448]\)}\end{tabular}\\[2pt]
GPT-OSS-20B & \begin{tabular}[t]{@{}c@{}}\(0.1621\)\\\ci{\([0.1313, 0.1941]\)}\end{tabular} & \begin{tabular}[t]{@{}c@{}}\(0.3172\)\\\ci{\([0.2752, 0.3595]\)}\end{tabular} & \begin{tabular}[t]{@{}c@{}}\(0.0466\)\\\ci{\([0.0308, 0.0636]\)}\end{tabular} & \begin{tabular}[t]{@{}c@{}}\(0.4224\)\\\ci{\([0.3681, 0.4775]\)}\end{tabular} & \begin{tabular}[t]{@{}c@{}}\({-}0.4155\)\\\ci{\([{-}0.4710, {-}0.3603]\)}\end{tabular}\\[2pt]
GPT-OSS-120B & \begin{tabular}[t]{@{}c@{}}\(0.2293\)\\\ci{\([0.1926, 0.2668]\)}\end{tabular} & \begin{tabular}[t]{@{}c@{}}\(0.3086\)\\\ci{\([0.2683, 0.3493]\)}\end{tabular} & \begin{tabular}[t]{@{}c@{}}\(0.1052\)\\\ci{\([0.0804, 0.1311]\)}\end{tabular} & \begin{tabular}[t]{@{}c@{}}\(0.0983\)\\\ci{\([0.0741, 0.1237]\)}\end{tabular} & \begin{tabular}[t]{@{}c@{}}\({-}0.0845\)\\\ci{\([{-}0.1096, {-}0.0603]\)}\end{tabular}\\[2pt]
\bottomrule
\end{tabular}
\endgroup

\end{table*}

\looseness=-1
In the full-step all-router fixed-receiver run of Appendix~\ref{app:wide-d},
switching the donor from correct to error on a fixed receiver changes only
$0.02$ (Nano) to $0.15$ (Qwen3.6) of verdicts, far fewer than patching itself.
On correct receivers, GPT-OSS-20B moves $0.10$ of verdicts to
\texttt{INCORRECT} and $0.04$ to \texttt{CORRECT}, and Gemma $0.09$ and $0.06$,
the categorical counterpart of their positive correct-receiver $D$. On error
receivers, by contrast, the accuracy change has the same sign and a similar size
under correct and erroneous donors in every checkpoint (for example $0.06$
versus $0.07$ in Lightning), so baseline-to-patched flips in that run are
dominated by patching itself rather than by donor role.

\subsection{Effects normalized by the baseline gap}

We normalize each panel by the baseline margin gap of the same eligible pairs
under the same margin, $R=\mathrm{mean}(\teffect)/\mathrm{mean}(g)$. By
Equation~\ref{eq:gap-compression}, $2\teffect=g-g_{\mathrm{after}}$ with
$g_{\mathrm{after}}=\marginm(x_{\error\leftarrow\clean})
-\marginm(x_{\clean\leftarrow\error})$, so $R=0.5$ means that the mean
post-transfer gap is zero and $R=1$ an equal-magnitude reversal. $R$ is a ratio
of means: it can be negative or exceed one, it is neither a mediation share nor
an explained variance, and it differs from dividing each receiver's effect by
its own distance to the decision boundary.

The full-step residual $R$ interval lies above $0.5$ in GPT-OSS-20B, Qwen3,
Qwen3.6, and Gemma, whereas $R$ is near $0.5$ in GLM-4.7-Flash and OLMoE and
near $0.25$ in Lightning, Nano, and GPT-OSS-120B
(Table~\ref{tab:normalized-effects}). Full-step
all-router $R$ runs from $0.17$ to $0.34$, reducing the mean gap to between 33\%
and 66\% of baseline, and single-layer router-logit $R$ stays within
$[-0.01,0.02]$ in all windows.

\begin{table*}[tbp]
\centering
\caption{Paired-donor effects normalized by the baseline margin gap of the same
pairs, $R=\mathrm{mean}(\teffect)/\mathrm{mean}(g)$, under the mean-logprob
margin. The last column gives the fraction of pairs with a non-positive baseline
gap.}
\label{tab:normalized-effects}
\begingroup
\renewcommand{\arraystretch}{1.15}
\tablestyle
\begin{tabular}{@{}lccccc@{}}
\toprule
Model & Mean gap \(g\) & \begin{tabular}[t]{@{}c@{}}\(R\), all-router\\(full step)\end{tabular} & \begin{tabular}[t]{@{}c@{}}\(R\), residual\\(full step)\end{tabular} & \begin{tabular}[t]{@{}c@{}}\(R\), residual\\(last token)\end{tabular} & \begin{tabular}[t]{@{}c@{}}Pairs with\\\(g\le 0\)\end{tabular} \\
\midrule
Qwen3.6 & \begin{tabular}[t]{@{}c@{}}\(1.5549\)\\\ci{\([1.3789, 1.7296]\)}\end{tabular} & \begin{tabular}[t]{@{}c@{}}\(0.2229\)\\\ci{\([0.1933, 0.2523]\)}\end{tabular} & \begin{tabular}[t]{@{}c@{}}\(0.8246\)\\\ci{\([0.7991, 0.8505]\)}\end{tabular} & \begin{tabular}[t]{@{}c@{}}\(0.0291\)\\\ci{\([0.0088, 0.0514]\)}\end{tabular} & \(0.1379\)\\[2pt]
Qwen3 & \begin{tabular}[t]{@{}c@{}}\(4.6818\)\\\ci{\([4.1758, 5.1846]\)}\end{tabular} & \begin{tabular}[t]{@{}c@{}}\(0.3365\)\\\ci{\([0.3057, 0.3666]\)}\end{tabular} & \begin{tabular}[t]{@{}c@{}}\(0.8982\)\\\ci{\([0.8786, 0.9170]\)}\end{tabular} & \begin{tabular}[t]{@{}c@{}}\(0.0892\)\\\ci{\([0.0737, 0.1050]\)}\end{tabular} & \(0.0897\)\\[2pt]
GLM-4.7-Flash & \begin{tabular}[t]{@{}c@{}}\(0.8147\)\\\ci{\([0.7125, 0.9175]\)}\end{tabular} & \begin{tabular}[t]{@{}c@{}}\(0.2839\)\\\ci{\([0.2481, 0.3179]\)}\end{tabular} & \begin{tabular}[t]{@{}c@{}}\(0.5177\)\\\ci{\([0.4857, 0.5496]\)}\end{tabular} & \begin{tabular}[t]{@{}c@{}}\(0.2856\)\\\ci{\([0.2566, 0.3149]\)}\end{tabular} & \(0.1586\)\\[2pt]
Lightning & \begin{tabular}[t]{@{}c@{}}\(0.8680\)\\\ci{\([0.7716, 0.9655]\)}\end{tabular} & \begin{tabular}[t]{@{}c@{}}\(0.2541\)\\\ci{\([0.2257, 0.2821]\)}\end{tabular} & \begin{tabular}[t]{@{}c@{}}\(0.2503\)\\\ci{\([0.2351, 0.2657]\)}\end{tabular} & \begin{tabular}[t]{@{}c@{}}\(0.1726\)\\\ci{\([0.1602, 0.1854]\)}\end{tabular} & \(0.1345\)\\[2pt]
Nano & \begin{tabular}[t]{@{}c@{}}\(0.5024\)\\\ci{\([0.4378, 0.5689]\)}\end{tabular} & \begin{tabular}[t]{@{}c@{}}\(0.2856\)\\\ci{\([0.2534, 0.3168]\)}\end{tabular} & \begin{tabular}[t]{@{}c@{}}\(0.2586\)\\\ci{\([0.2412, 0.2764]\)}\end{tabular} & \begin{tabular}[t]{@{}c@{}}\(0.1726\)\\\ci{\([0.1612, 0.1845]\)}\end{tabular} & \(0.1655\)\\[2pt]
Gemma 4 & \begin{tabular}[t]{@{}c@{}}\(7.0863\)\\\ci{\([6.3870, 7.7836]\)}\end{tabular} & \begin{tabular}[t]{@{}c@{}}\(0.1705\)\\\ci{\([0.1388, 0.2025]\)}\end{tabular} & \begin{tabular}[t]{@{}c@{}}\(0.6872\)\\\ci{\([0.6520, 0.7226]\)}\end{tabular} & \begin{tabular}[t]{@{}c@{}}\(0.1171\)\\\ci{\([0.0891, 0.1478]\)}\end{tabular} & \(0.0931\)\\[2pt]
OLMoE & \begin{tabular}[t]{@{}c@{}}\(0.6056\)\\\ci{\([0.4427, 0.7738]\)}\end{tabular} & \begin{tabular}[t]{@{}c@{}}\(0.3059\)\\\ci{\([0.2277, 0.3846]\)}\end{tabular} & \begin{tabular}[t]{@{}c@{}}\(0.5356\)\\\ci{\([0.4726, 0.6084]\)}\end{tabular} & \begin{tabular}[t]{@{}c@{}}\(0.1577\)\\\ci{\([0.1005, 0.2200]\)}\end{tabular} & \(0.3966\)\\[2pt]
GPT-OSS-20B & \begin{tabular}[t]{@{}c@{}}\(1.2942\)\\\ci{\([1.1356, 1.4525]\)}\end{tabular} & \begin{tabular}[t]{@{}c@{}}\(0.2293\)\\\ci{\([0.1932, 0.2663]\)}\end{tabular} & \begin{tabular}[t]{@{}c@{}}\(0.9040\)\\\ci{\([0.8819, 0.9268]\)}\end{tabular} & \begin{tabular}[t]{@{}c@{}}\(0.1134\)\\\ci{\([0.0927, 0.1357]\)}\end{tabular} & \(0.1310\)\\[2pt]
GPT-OSS-120B & \begin{tabular}[t]{@{}c@{}}\(1.3498\)\\\ci{\([1.1818, 1.5193]\)}\end{tabular} & \begin{tabular}[t]{@{}c@{}}\(0.3173\)\\\ci{\([0.2747, 0.3589]\)}\end{tabular} & \begin{tabular}[t]{@{}c@{}}\(0.2371\)\\\ci{\([0.2156, 0.2602]\)}\end{tabular} & \begin{tabular}[t]{@{}c@{}}\(0.1550\)\\\ci{\([0.1386, 0.1724]\)}\end{tabular} & \(0.1931\)\\[2pt]
\bottomrule
\end{tabular}
\endgroup

\end{table*}

Within a checkpoint and margin definition, normalization preserves interface
orderings when panels share an eligible-pair set and a positive mean baseline
gap; across windows or controls with different eligible sets, the denominator
changes and orderings need not be preserved. On the residual interface, matched
external donors give essentially the same $R$ as paired donors ($0.25$ versus
$0.25$ in Lightning, $0.26$ versus $0.26$ in Nano, and $0.25$ versus $0.24$ in
GPT-OSS-120B), so normalization does not
change the donor-generic character of that effect. All 1,152 panels have
denominators whose intervals exclude zero. OLMoE has non-positive baseline gaps
in 40\% of pairs, against at most 19\% elsewhere, so its $R$ should not be read
as a per-pair recovery fraction.

\subsection{Unpatched scores of the full-step all-router run}

The unpatched candidate scores of the run in Appendix~\ref{app:wide-d} reproduce
the window-run scores receiver by receiver for Qwen3.6, Qwen3, OLMoE,
GPT-OSS-20B, and GPT-OSS-120B, but not for GLM-4.7-Flash, Lightning, Nano, and
Gemma. Recorded inputs, tokenization, pair identities, and receiver roles agree
across the two runs for all nine checkpoints, and the source of the score
discrepancies remains unresolved; unpatched accuracy differs by at most $0.006$.
Table~\ref{tab:verification-baseline} reports the window-run scores, and each
run retains its own baseline. The fixed-receiver $D$ estimates of
Appendix~\ref{app:wide-d} compare patched scores within that run and do not use
the window-run baseline.

\section{Single-Layer Fixed-Receiver Results}
\label{app:local-d}

Table~\ref{tab:abc-d-all-interfaces} reports last-token fixed-receiver $D$ at
all three local interfaces and both receiver types. Every single-router
$D_{\mathrm{common}}$ interval includes zero, whereas residual $D$ is positive
on both receiver types in eight checkpoints.

\begin{table*}[tbp]
\centering
\caption{Last-token fixed-receiver $D$ at all three local interfaces.
$D_{\mathrm{common}}$ averages the two receiver types within pair. Qwen3 uses
L36 for this experiment and L24 for the token-window experiment. A substantial
part of OLMoE's residual $D$ is associated with surface content
(Appendix~\ref{app:surface-routing}).}
\label{tab:abc-d-all-interfaces}
\begingroup
\renewcommand{\arraystretch}{1.15}
\tablestyle
\begin{tabular}{@{}lccc@{}}
\toprule
Model & \(D_{\mathrm{correct}}\) & \(D_{\mathrm{error}}\) & \(D_{\mathrm{common}}\) \\
\midrule
\multicolumn{4}{l}{\textit{Router}} \\[1pt]
Qwen3.6 & \(\phantom{{-}}.0015\)\hspace{0.25em}\ci{\([{-}.0056, \phantom{1}.0093]\)} & \({-}.0035\)\hspace{0.25em}\ci{\([{-}.0168, .0091]\)} & \({-}.0010\)\hspace{0.25em}\ci{\([{-}.0085, .0062]\)}\\[1pt]
Qwen3 & \({-}.0054\)\hspace{0.25em}\ci{\([{-}.0157, \phantom{1}.0049]\)} & \({-}.0091\)\hspace{0.25em}\ci{\([{-}.0197, .0015]\)} & \({-}.0073\)\hspace{0.25em}\ci{\([{-}.0151, .0006]\)}\\[1pt]
GLM-4.7-Flash & \({-}.0005\)\hspace{0.25em}\ci{\([{-}.0109, \phantom{1}.0100]\)} & \(\phantom{{-}}.0022\)\hspace{0.25em}\ci{\([{-}.0084, .0130]\)} & \(\phantom{{-}}.0009\)\hspace{0.25em}\ci{\([{-}.0062, .0080]\)}\\[1pt]
Lightning & \(\phantom{{-}}.0017\)\hspace{0.25em}\ci{\([{-}.0026, \phantom{1}.0061]\)} & \({-}.0018\)\hspace{0.25em}\ci{\([{-}.0072, .0037]\)} & \({-}.0001\)\hspace{0.25em}\ci{\([{-}.0035, .0034]\)}\\[1pt]
Nano & \(\phantom{{-}}.0018\)\hspace{0.25em}\ci{\([{-}.0012, \phantom{1}.0047]\)} & \({-}.0008\)\hspace{0.25em}\ci{\([{-}.0038, .0022]\)} & \(\phantom{{-}}.0005\)\hspace{0.25em}\ci{\([{-}.0017, .0026]\)}\\[1pt]
Gemma 4 & \({-}.0106\)\hspace{0.25em}\ci{\([{-}.0328, \phantom{1}.0111]\)} & \({-}.0016\)\hspace{0.25em}\ci{\([{-}.0202, .0180]\)} & \({-}.0061\)\hspace{0.25em}\ci{\([{-}.0208, .0087]\)}\\[1pt]
OLMoE & \(\phantom{{-}}.0061\)\hspace{0.25em}\ci{\([{-}.0011, \phantom{1}.0132]\)} & \(\phantom{{-}}.0039\)\hspace{0.25em}\ci{\([{-}.0049, .0127]\)} & \(\phantom{{-}}.0050\)\hspace{0.25em}\ci{\([{-}.0011, .0110]\)}\\[1pt]
GPT-OSS-20B & \({-}.0004\)\hspace{0.25em}\ci{\([{-}.0049, \phantom{1}.0042]\)} & \({-}.0029\)\hspace{0.25em}\ci{\([{-}.0103, .0043]\)} & \({-}.0017\)\hspace{0.25em}\ci{\([{-}.0060, .0027]\)}\\[1pt]
GPT-OSS-120B & \(\phantom{{-}}.0031\)\hspace{0.25em}\ci{\([{-}.0093, \phantom{1}.0156]\)} & \(\phantom{{-}}.0092\)\hspace{0.25em}\ci{\([{-}.0044, .0228]\)} & \(\phantom{{-}}.0061\)\hspace{0.25em}\ci{\([{-}.0025, .0150]\)}\\[1pt]
\midrule
\multicolumn{4}{l}{\textit{MoE output}} \\[1pt]
Qwen3.6 & \(\phantom{{-}}.0009\)\hspace{0.25em}\ci{\([{-}.0063, \phantom{1}.0080]\)} & \({-}.0115\)\hspace{0.25em}\ci{\([{-}.0240, .0003]\)} & \({-}.0053\)\hspace{0.25em}\ci{\([{-}.0120, .0010]\)}\\[1pt]
Qwen3 & \(\phantom{{-}}.0498\)\hspace{0.25em}\ci{\([\phantom{{-}}.0321, \phantom{1}.0676]\)} & \(\phantom{{-}}.0540\)\hspace{0.25em}\ci{\([\phantom{{-}}.0305, .0771]\)} & \(\phantom{{-}}.0519\)\hspace{0.25em}\ci{\([\phantom{{-}}.0339, .0698]\)}\\[1pt]
GLM-4.7-Flash & \(\phantom{{-}}.0802\)\hspace{0.25em}\ci{\([\phantom{{-}}.0598, \phantom{1}.1006]\)} & \(\phantom{{-}}.0482\)\hspace{0.25em}\ci{\([\phantom{{-}}.0316, .0647]\)} & \(\phantom{{-}}.0642\)\hspace{0.25em}\ci{\([\phantom{{-}}.0475, .0806]\)}\\[1pt]
Lightning & \(\phantom{{-}}.0006\)\hspace{0.25em}\ci{\([{-}.0039, \phantom{1}.0052]\)} & \({-}.0012\)\hspace{0.25em}\ci{\([{-}.0064, .0040]\)} & \({-}.0003\)\hspace{0.25em}\ci{\([{-}.0041, .0036]\)}\\[1pt]
Nano & \(\phantom{{-}}.0018\)\hspace{0.25em}\ci{\([{-}.0015, \phantom{1}.0051]\)} & \({-}.0010\)\hspace{0.25em}\ci{\([{-}.0048, .0029]\)} & \(\phantom{{-}}.0004\)\hspace{0.25em}\ci{\([{-}.0024, .0032]\)}\\[1pt]
Gemma 4 & \(\phantom{{-}}.1430\)\hspace{0.25em}\ci{\([\phantom{{-}}.0727, \phantom{1}.2295]\)} & \(\phantom{{-}}.0946\)\hspace{0.25em}\ci{\([\phantom{{-}}.0459, .1458]\)} & \(\phantom{{-}}.1188\)\hspace{0.25em}\ci{\([\phantom{{-}}.0681, .1793]\)}\\[1pt]
OLMoE & \(\phantom{{-}}.0402\)\hspace{0.25em}\ci{\([\phantom{{-}}.0250, \phantom{1}.0556]\)} & \(\phantom{{-}}.0296\)\hspace{0.25em}\ci{\([\phantom{{-}}.0123, .0472]\)} & \(\phantom{{-}}.0349\)\hspace{0.25em}\ci{\([\phantom{{-}}.0211, .0487]\)}\\[1pt]
GPT-OSS-20B & \(\phantom{{-}}.0031\)\hspace{0.25em}\ci{\([{-}.0094, \phantom{1}.0154]\)} & \({-}.0059\)\hspace{0.25em}\ci{\([{-}.0167, .0048]\)} & \({-}.0014\)\hspace{0.25em}\ci{\([{-}.0108, .0079]\)}\\[1pt]
GPT-OSS-120B & \(\phantom{{-}}.0433\)\hspace{0.25em}\ci{\([\phantom{{-}}.0290, \phantom{1}.0581]\)} & \(\phantom{{-}}.0439\)\hspace{0.25em}\ci{\([\phantom{{-}}.0268, .0617]\)} & \(\phantom{{-}}.0436\)\hspace{0.25em}\ci{\([\phantom{{-}}.0312, .0564]\)}\\[1pt]
\midrule
\multicolumn{4}{l}{\textit{Residual}} \\[1pt]
Qwen3.6 & \({-}.0096\)\hspace{0.25em}\ci{\([{-}.0308, \phantom{1}.0130]\)} & \(\phantom{{-}}.0356\)\hspace{0.25em}\ci{\([\phantom{{-}}.0101, .0625]\)} & \(\phantom{{-}}.0130\)\hspace{0.25em}\ci{\([{-}.0054, .0327]\)}\\[1pt]
Qwen3 & \(\phantom{{-}}.7282\)\hspace{0.25em}\ci{\([\phantom{{-}}.6313, \phantom{1}.8269]\)} & \(\phantom{{-}}.5921\)\hspace{0.25em}\ci{\([\phantom{{-}}.4989, .6879]\)} & \(\phantom{{-}}.6601\)\hspace{0.25em}\ci{\([\phantom{{-}}.5712, .7509]\)}\\[1pt]
GLM-4.7-Flash & \(\phantom{{-}}.1989\)\hspace{0.25em}\ci{\([\phantom{{-}}.1594, \phantom{1}.2404]\)} & \(\phantom{{-}}.1371\)\hspace{0.25em}\ci{\([\phantom{{-}}.1077, .1673]\)} & \(\phantom{{-}}.1680\)\hspace{0.25em}\ci{\([\phantom{{-}}.1353, .2019]\)}\\[1pt]
Lightning & \(\phantom{{-}}.1492\)\hspace{0.25em}\ci{\([\phantom{{-}}.1285, \phantom{1}.1704]\)} & \(\phantom{{-}}.1125\)\hspace{0.25em}\ci{\([\phantom{{-}}.0953, .1304]\)} & \(\phantom{{-}}.1309\)\hspace{0.25em}\ci{\([\phantom{{-}}.1127, .1495]\)}\\[1pt]
Nano & \(\phantom{{-}}.0793\)\hspace{0.25em}\ci{\([\phantom{{-}}.0680, \phantom{1}.0907]\)} & \(\phantom{{-}}.0624\)\hspace{0.25em}\ci{\([\phantom{{-}}.0525, .0726]\)} & \(\phantom{{-}}.0709\)\hspace{0.25em}\ci{\([\phantom{{-}}.0608, .0811]\)}\\[1pt]
Gemma 4 & \(\phantom{{-}}\llap{\(1\)}.1705\)\hspace{0.25em}\ci{\([\phantom{{-}}.8775, 1.4791]\)} & \(\phantom{{-}}.2586\)\hspace{0.25em}\ci{\([\phantom{{-}}.1150, .4163]\)} & \(\phantom{{-}}.7145\)\hspace{0.25em}\ci{\([\phantom{{-}}.5321, .9069]\)}\\[1pt]
OLMoE & \(\phantom{{-}}.1368\)\hspace{0.25em}\ci{\([\phantom{{-}}.0859, \phantom{1}.1925]\)} & \(\phantom{{-}}.1586\)\hspace{0.25em}\ci{\([\phantom{{-}}.1049, .2169]\)} & \(\phantom{{-}}.1477\)\hspace{0.25em}\ci{\([\phantom{{-}}.0966, .2038]\)}\\[1pt]
GPT-OSS-20B & \(\phantom{{-}}.1352\)\hspace{0.25em}\ci{\([\phantom{{-}}.1016, \phantom{1}.1696]\)} & \(\phantom{{-}}.1019\)\hspace{0.25em}\ci{\([\phantom{{-}}.0726, .1323]\)} & \(\phantom{{-}}.1186\)\hspace{0.25em}\ci{\([\phantom{{-}}.0909, .1473]\)}\\[1pt]
GPT-OSS-120B & \(\phantom{{-}}.2262\)\hspace{0.25em}\ci{\([\phantom{{-}}.1895, \phantom{1}.2645]\)} & \(\phantom{{-}}.1654\)\hspace{0.25em}\ci{\([\phantom{{-}}.1342, .1977]\)} & \(\phantom{{-}}.1958\)\hspace{0.25em}\ci{\([\phantom{{-}}.1659, .2271]\)}\\[1pt]
\bottomrule
\end{tabular}
\endgroup

\end{table*}

\paragraph{Receiver dependence.}
Qwen3.6 is the only checkpoint whose local residual intervention lacks a
consistent positive direction across receiver types. Its $D_{\clean}$ is
$-0.0096\;[-0.0308,0.0130]$, whereas $D_{\error}$ is $0.0356\;[0.0101,0.0625]$,
and their direct difference, $-0.0452\;[-0.0748,-0.0156]$, establishes receiver
dependence. Its $D_{\mathrm{common}}$ has median $-0.0008$ and 10\%-trimmed
mean $0.0022$, so the positive mean does not reflect a uniform pairwise
direction. The remaining eight residual panels have positive intervals for both
receiver types, although their magnitudes need not match; for example, the
correct-minus-error receiver differences for Qwen3 and GLM are
$0.136\;[0.069,0.205]$ and $0.062\;[0.038,0.087]$.

\section{Full-Step All-Router Fixed-Receiver Analysis}
\label{app:wide-d}

\subsection{Experimental design and primary analysis}

This experiment extends the fixed-receiver contrast in Equation~\ref{eq:d} to
every adapter-defined routed layer listed in Table~\ref{tab:abc-anchors} and
every token in the evaluated step, with $\alpha=1$. All nine checkpoints use the
same 290 receiver pairs, 235 source components, and a fixed assignment to 204
distinct external donor pairs, matched as in
Appendix~\ref{app:nine-model-protocols} after excluding the receiver pair and
its source component. For each receiver, the correct and error members of the
matched donor pair separately replace router logits across all routed layers
and the full step, with donor and receiver spans aligned by the monotone
mapping of Appendix~\ref{app:nine-model-protocols}. Each checkpoint retains its
native BF16 or MXFP4 execution and native expert dispatch.

The primary endpoint remains the mean-logprob verification margin of
Equation~\ref{eq:margin}. For each receiver pair, the four donor/receiver
combinations yield $D_{\clean}$, $D_{\error}$, and their within-pair average
$D_{\mathrm{common}}$. All four intervention scores and identity controls are
generated in the same loaded model invocation, and all 290 pairs are complete
for every checkpoint, so no values are imputed. A positive
$D_{\mathrm{common}}$ indicates an average correctness-dependent effect across
the two receiver types; it does not require both receiver-specific effects, or
every pair effect, to be positive.

Resampling follows Appendix~\ref{app:statistics} with seed 20260908. Both
receiver roles share each draw, so $D_{\mathrm{common}}$ has $N=290$, not 580,
and donors are neither rematched nor resampled independently.
Table~\ref{tab:wide-d-primary} reports the primary results and
Table~\ref{tab:wide-d-total} the scoring sensitivity. Counts of intervals
excluding zero should not be read as estimates of a model-population success
rate.

\begin{table*}[tbp]
\centering
\caption{Primary full-step all-router fixed-receiver effects, with the same
fixed assignment to 204 external donor pairs in every checkpoint. Entries are
mean-logprob estimates at $\alpha=1$. $D_{\mathrm{common}}$ averages the two
receiver types within pair. Checkpoint-specific margin scales should not be
interpreted as ability rankings.}
\label{tab:wide-d-primary}
\begingroup
\renewcommand{\arraystretch}{1.08}%
\tablestyle
\begin{tabular}{@{}lccc@{}}
\toprule
Model & \(D_{\mathrm{correct}}\) & \(D_{\mathrm{error}}\) & \(D_{\mathrm{common}}\) \\
\midrule
Qwen3.6 & \begin{tabular}[t]{@{}c@{}}\(\phantom{{-}0}\llap{\(0\)}.0198\)\\\ci{\([{-}0.0199, 0.0595]\)}\end{tabular} & \begin{tabular}[t]{@{}c@{}}\(\phantom{{-}0}\llap{\(0\)}.0012\)\\\ci{\([{-}0.0616, \phantom{{-}0}\llap{\(0\)}.0644]\)}\end{tabular} & \begin{tabular}[t]{@{}c@{}}\(\phantom{{-}0}\llap{\(0\)}.0105\)\\\ci{\([{-}0.0282, 0.0486]\)}\end{tabular}\\[2pt]
Qwen3 & \begin{tabular}[t]{@{}c@{}}\({-}0.0954\)\\\ci{\([{-}0.2095, 0.0192]\)}\end{tabular} & \begin{tabular}[t]{@{}c@{}}\({-}0.0738\)\\\ci{\([{-}0.2102, \phantom{{-}0}\llap{\(0\)}.0589]\)}\end{tabular} & \begin{tabular}[t]{@{}c@{}}\({-}0.0846\)\\\ci{\([{-}0.1872, 0.0166]\)}\end{tabular}\\[2pt]
GLM-4.7-Flash & \begin{tabular}[t]{@{}c@{}}\(\phantom{{-}0}\llap{\(0\)}.0041\)\\\ci{\([{-}0.0273, 0.0369]\)}\end{tabular} & \begin{tabular}[t]{@{}c@{}}\({-}0.0292\)\\\ci{\([{-}0.0572, {-}0.0011]\)}\end{tabular} & \begin{tabular}[t]{@{}c@{}}\({-}0.0125\)\\\ci{\([{-}0.0363, 0.0110]\)}\end{tabular}\\[2pt]
Lightning & \begin{tabular}[t]{@{}c@{}}\(\phantom{{-}0}\llap{\(0\)}.0213\)\\\ci{\([{-}0.0005, 0.0430]\)}\end{tabular} & \begin{tabular}[t]{@{}c@{}}\(\phantom{{-}0}\llap{\(0\)}.0043\)\\\ci{\([{-}0.0197, \phantom{{-}0}\llap{\(0\)}.0280]\)}\end{tabular} & \begin{tabular}[t]{@{}c@{}}\(\phantom{{-}0}\llap{\(0\)}.0128\)\\\ci{\([{-}0.0050, 0.0304]\)}\end{tabular}\\[2pt]
Nano & \begin{tabular}[t]{@{}c@{}}\(\phantom{{-}0}\llap{\(0\)}.0032\)\\\ci{\([{-}0.0109, 0.0171]\)}\end{tabular} & \begin{tabular}[t]{@{}c@{}}\({-}0.0094\)\\\ci{\([{-}0.0219, \phantom{{-}0}\llap{\(0\)}.0028]\)}\end{tabular} & \begin{tabular}[t]{@{}c@{}}\({-}0.0031\)\\\ci{\([{-}0.0134, 0.0069]\)}\end{tabular}\\[2pt]
Gemma 4 & \begin{tabular}[t]{@{}c@{}}\(\phantom{{-}0}\llap{\(0\)}.3875\)\\\ci{\([\phantom{{-}0}\llap{\(0\)}.0762, 0.7002]\)}\end{tabular} & \begin{tabular}[t]{@{}c@{}}\(\phantom{{-}0}\llap{\(0\)}.0540\)\\\ci{\([{-}0.2159, \phantom{{-}0}\llap{\(0\)}.3335]\)}\end{tabular} & \begin{tabular}[t]{@{}c@{}}\(\phantom{{-}0}\llap{\(0\)}.2207\)\\\ci{\([{-}0.0114, 0.4545]\)}\end{tabular}\\[2pt]
OLMoE & \begin{tabular}[t]{@{}c@{}}\({-}0.0088\)\\\ci{\([{-}0.0609, 0.0435]\)}\end{tabular} & \begin{tabular}[t]{@{}c@{}}\(\phantom{{-}0}\llap{\(0\)}.0415\)\\\ci{\([{-}0.0435, \phantom{{-}0}\llap{\(0\)}.1270]\)}\end{tabular} & \begin{tabular}[t]{@{}c@{}}\(\phantom{{-}0}\llap{\(0\)}.0164\)\\\ci{\([{-}0.0376, 0.0713]\)}\end{tabular}\\[2pt]
GPT-OSS-20B & \begin{tabular}[t]{@{}c@{}}\(\phantom{{-}0}\llap{\(0\)}.1137\)\\\ci{\([\phantom{{-}0}\llap{\(0\)}.0480, 0.1803]\)}\end{tabular} & \begin{tabular}[t]{@{}c@{}}\(\phantom{{-}0}\llap{\(0\)}.0262\)\\\ci{\([{-}0.0335, \phantom{{-}0}\llap{\(0\)}.0871]\)}\end{tabular} & \begin{tabular}[t]{@{}c@{}}\(\phantom{{-}0}\llap{\(0\)}.0699\)\\\ci{\([\phantom{{-}0}\llap{\(0\)}.0181, 0.1226]\)}\end{tabular}\\[2pt]
GPT-OSS-120B & \begin{tabular}[t]{@{}c@{}}\({-}0.0032\)\\\ci{\([{-}0.0489, 0.0435]\)}\end{tabular} & \begin{tabular}[t]{@{}c@{}}\({-}0.0071\)\\\ci{\([{-}0.0539, \phantom{{-}0}\llap{\(0\)}.0396]\)}\end{tabular} & \begin{tabular}[t]{@{}c@{}}\({-}0.0051\)\\\ci{\([{-}0.0393, 0.0290]\)}\end{tabular}\\[2pt]
\bottomrule
\end{tabular}%
\endgroup

\end{table*}

\subsection{Candidate weighting sensitivity}

Total log probability uses the same candidate scores as the primary endpoint
but weights candidates by their token counts, so it is a scoring sensitivity
rather than an independent replication. Gemma illustrates the distinction. All
1,160 four-cell scoring rows contain two target tokens for \texttt{CORRECT} and
three for \texttt{INCORRECT}. Let $A$ denote the error-minus-correct donor
change in the \texttt{INCORRECT} candidate's mean log probability, averaged
first over the two receivers and then over pairs, and let $B$ denote the
corresponding change for \texttt{CORRECT}. Then
\begin{equation}
D_{\mathrm{common}}^{\mathrm{mean}}=A-B,\qquad
D_{\mathrm{common}}^{\mathrm{total}}=3A-2B.
\label{eq:wide-d-gemma-weighting}
\end{equation}

\looseness=-1
Here $A=0.107$ and $B=-0.113$. The primary mean-logprob result is
$0.221\;[-0.011,0.455]$, whereas the total-logprob result is
$0.549\;[0.003,1.102]$. The secondary analysis verifies this reweighting for
every pair and every bootstrap draw, with maximum absolute residual
$1.78\times10^{-15}$. The change is not a uniform positive rescaling, which would
preserve whether an interval contains zero. Gemma's correct-receiver interval is
positive under both endpoints, while its primary $D_{\mathrm{common}}$ interval
includes zero. GLM's error-receiver interval is slightly negative under mean
scoring and includes zero under total scoring. Mean scoring remains the primary
endpoint for all checkpoints.

\begin{table*}[tbp]
\centering
\caption{Total-logprob sensitivity for the same full-step all-router scores,
pairs, donor assignments, and bootstrap draws as
Table~\ref{tab:wide-d-primary}. Candidate totals change the relative weighting
of verdict strings when their token counts differ. These results do not replace
the primary mean-logprob endpoint and are not an independent replication.}
\label{tab:wide-d-total}
\begingroup
\renewcommand{\arraystretch}{1.08}%
\tablestyle
\begin{tabular}{@{}lccc@{}}
\toprule
Model & \(D_{\mathrm{correct}}\) & \(D_{\mathrm{error}}\) & \(D_{\mathrm{common}}\) \\
\midrule
Qwen3.6 & \begin{tabular}[t]{@{}c@{}}\(\phantom{{-}0}\llap{\(0\)}.0477\)\\\ci{\([{-}0.0541, 0.1509]\)}\end{tabular} & \begin{tabular}[t]{@{}c@{}}\({-}0.0055\)\\\ci{\([{-}0.1502, 0.1411]\)}\end{tabular} & \begin{tabular}[t]{@{}c@{}}\(\phantom{{-}0}\llap{\(0\)}.0211\)\\\ci{\([{-}0.0710, 0.1118]\)}\end{tabular}\\[2pt]
Qwen3 & \begin{tabular}[t]{@{}c@{}}\({-}0.2638\)\\\ci{\([{-}0.5815, 0.0554]\)}\end{tabular} & \begin{tabular}[t]{@{}c@{}}\({-}0.1974\)\\\ci{\([{-}0.5702, 0.1644]\)}\end{tabular} & \begin{tabular}[t]{@{}c@{}}\({-}0.2306\)\\\ci{\([{-}0.5154, 0.0491]\)}\end{tabular}\\[2pt]
GLM-4.7-Flash & \begin{tabular}[t]{@{}c@{}}\(\phantom{{-}0}\llap{\(0\)}.0147\)\\\ci{\([{-}0.0753, 0.1090]\)}\end{tabular} & \begin{tabular}[t]{@{}c@{}}\({-}0.0774\)\\\ci{\([{-}0.1552, 0.0009]\)}\end{tabular} & \begin{tabular}[t]{@{}c@{}}\({-}0.0313\)\\\ci{\([{-}0.0986, 0.0356]\)}\end{tabular}\\[2pt]
Lightning & \begin{tabular}[t]{@{}c@{}}\(\phantom{{-}0}\llap{\(0\)}.0717\)\\\ci{\([{-}0.0052, 0.1485]\)}\end{tabular} & \begin{tabular}[t]{@{}c@{}}\(\phantom{{-}0}\llap{\(0\)}.0040\)\\\ci{\([{-}0.0775, 0.0838]\)}\end{tabular} & \begin{tabular}[t]{@{}c@{}}\(\phantom{{-}0}\llap{\(0\)}.0379\)\\\ci{\([{-}0.0239, 0.0990]\)}\end{tabular}\\[2pt]
Nano & \begin{tabular}[t]{@{}c@{}}\(\phantom{{-}0}\llap{\(0\)}.0090\)\\\ci{\([{-}0.0375, 0.0549]\)}\end{tabular} & \begin{tabular}[t]{@{}c@{}}\({-}0.0307\)\\\ci{\([{-}0.0702, 0.0082]\)}\end{tabular} & \begin{tabular}[t]{@{}c@{}}\({-}0.0108\)\\\ci{\([{-}0.0442, 0.0217]\)}\end{tabular}\\[2pt]
Gemma 4 & \begin{tabular}[t]{@{}c@{}}\(\phantom{{-}0}\llap{\(0\)}.9582\)\\\ci{\([\phantom{{-}0}\llap{\(0\)}.2172, 1.7054]\)}\end{tabular} & \begin{tabular}[t]{@{}c@{}}\(\phantom{{-}0}\llap{\(0\)}.1395\)\\\ci{\([{-}0.4944, 0.7933]\)}\end{tabular} & \begin{tabular}[t]{@{}c@{}}\(\phantom{{-}0}\llap{\(0\)}.5488\)\\\ci{\([\phantom{{-}0}\llap{\(0\)}.0034, 1.1020]\)}\end{tabular}\\[2pt]
OLMoE & \begin{tabular}[t]{@{}c@{}}\({-}0.0254\)\\\ci{\([{-}0.1765, 0.1265]\)}\end{tabular} & \begin{tabular}[t]{@{}c@{}}\(\phantom{{-}0}\llap{\(0\)}.1043\)\\\ci{\([{-}0.1255, 0.3361]\)}\end{tabular} & \begin{tabular}[t]{@{}c@{}}\(\phantom{{-}0}\llap{\(0\)}.0394\)\\\ci{\([{-}0.1092, 0.1911]\)}\end{tabular}\\[2pt]
GPT-OSS-20B & \begin{tabular}[t]{@{}c@{}}\(\phantom{{-}0}\llap{\(0\)}.2588\)\\\ci{\([\phantom{{-}0}\llap{\(0\)}.1017, 0.4193]\)}\end{tabular} & \begin{tabular}[t]{@{}c@{}}\(\phantom{{-}0}\llap{\(0\)}.0719\)\\\ci{\([{-}0.0600, 0.2074]\)}\end{tabular} & \begin{tabular}[t]{@{}c@{}}\(\phantom{{-}0}\llap{\(0\)}.1653\)\\\ci{\([\phantom{{-}0}\llap{\(0\)}.0463, 0.2868]\)}\end{tabular}\\[2pt]
GPT-OSS-120B & \begin{tabular}[t]{@{}c@{}}\(\phantom{{-}0}\llap{\(0\)}.0136\)\\\ci{\([{-}0.0968, 0.1269]\)}\end{tabular} & \begin{tabular}[t]{@{}c@{}}\({-}0.0245\)\\\ci{\([{-}0.1340, 0.0832]\)}\end{tabular} & \begin{tabular}[t]{@{}c@{}}\({-}0.0055\)\\\ci{\([{-}0.0859, 0.0749]\)}\end{tabular}\\[2pt]
\bottomrule
\end{tabular}%
\endgroup

\end{table*}

\subsection{Post hoc direct contrasts and input identity}

After inspecting the nine-checkpoint results, we computed two classes of
secondary direct contrast from the saved four-cell scores:
$D_{\clean}-D_{\error}$ within each checkpoint, and GPT-OSS-20B minus
GPT-OSS-120B in $D_{\mathrm{common}}$. Each difference is formed within receiver
pair before the same resampling, jointly resampling both receiver types or both
checkpoints. Table~\ref{tab:wide-d-direct} reports these contrasts.
GPT-OSS-20B has $D_{\clean}-D_{\error}=0.087\;[0.016,0.160]$. Gemma and GLM have
direct receiver-difference intervals that include zero, so the different
zero-crossing behavior of their marginal intervals does not establish a
receiver-type difference; conversely, an interval containing zero does not
establish equivalence.

All nine checkpoints share receiver sample identities, source components, and
fixed donor assignments. For the two GPT-OSS checkpoints, the secondary
analysis additionally verifies all 1,160 receiver/donor identity occurrences and
2,320 candidate-conditioned complete inputs, checking saved rendered-prompt
hashes, input formats, step spans, sample/source identities, target IDs and
positions, tensor shapes, and hashes recomputed from the complete
\texttt{input\_ids} arrays. Both checkpoints produce identical hashes across
the entire input projection (checksums are archived in the repository
manifests). Although the archived identity records do not separately store
rendered prompt text, matching whole-prompt hashes and freshly hashed complete
token arrays establish identity of the full inputs used in this comparison.
This check applies only to the full-step fixed-receiver GPT comparison and does
not establish identity with earlier $T$, local $D$, or representation-probe
experiments.

The paired GPT-OSS-20B-minus-120B contrast is $0.075\;[0.021,0.130]$. It
compares two fixed checkpoints under identical inputs and donor assignments,
but training, checkpoint identity, expert count, and routed-layer coverage still
differ, so it does not isolate parameter count. Similarly, comparing an
unresolved local-router interval with a positive full-step all-router interval
is not a direct test of expanding coverage, and full-step router and local
residual interventions also differ in layer and token coverage. Neither
comparison identifies an equal-coverage interface difference, a directional
fraction $D/T$, or a compression component $T-D$.

\begin{table*}[tbp]
\centering
\caption{Post hoc mean-logprob contrasts for full-step all-router $D$.
Receiver-type differences are correct minus error within checkpoint; the
checkpoint difference is GPT-OSS 20B minus 120B in $D_{\mathrm{common}}$.
Differences are formed within pair and jointly resampled over the same 235
receiver source components with fixed donors. The GPT checkpoint contrast uses
verified identical inputs but does not isolate parameter count.}
\label{tab:wide-d-direct}
\begingroup
\renewcommand{\arraystretch}{1.08}%
\tablestyle
\begin{tabular}{@{}lccc@{}}
\toprule
Model / checkpoint contrast & Estimate & 95\% CI & Bootstrap SE \\
\midrule
\multicolumn{4}{l}{\textit{Receiver-role difference: }\(D_{\mathrm{correct}}-D_{\mathrm{error}}\)}\\[3pt]
Qwen3.6 & \(\phantom{{-}0}\llap{\(0\)}.0185\) & \ci{\([{-}0.0545, 0.0897]\)} & \(0.0369\)\\[3pt]
Qwen3 & \({-}0.0216\) & \ci{\([{-}0.1624, 0.1248]\)} & \(0.0733\)\\[3pt]
GLM-4.7-Flash & \(\phantom{{-}0}\llap{\(0\)}.0332\) & \ci{\([{-}0.0033, 0.0712]\)} & \(0.0190\)\\[3pt]
Lightning & \(\phantom{{-}0}\llap{\(0\)}.0170\) & \ci{\([{-}0.0115, 0.0461]\)} & \(0.0147\)\\[3pt]
Nano & \(\phantom{{-}0}\llap{\(0\)}.0126\) & \ci{\([{-}0.0041, 0.0295]\)} & \(0.0086\)\\[3pt]
Gemma 4 & \(\phantom{{-}0}\llap{\(0\)}.3335\) & \ci{\([{-}0.0215, 0.6877]\)} & \(0.1804\)\\[3pt]
OLMoE & \({-}0.0503\) & \ci{\([{-}0.1387, 0.0406]\)} & \(0.0459\)\\[3pt]
GPT-OSS-20B & \(\phantom{{-}0}\llap{\(0\)}.0874\) & \ci{\([\phantom{{-}0}\llap{\(0\)}.0158, 0.1597]\)} & \(0.0368\)\\[3pt]
GPT-OSS-120B & \(\phantom{{-}0}\llap{\(0\)}.0039\) & \ci{\([{-}0.0589, 0.0673]\)} & \(0.0324\)\\[3pt]
\midrule
\multicolumn{4}{l}{\textit{Checkpoint difference in }\(D_{\mathrm{common}}\)}\\[3pt]
\begin{tabular}[c]{@{}l@{}}GPT-OSS-20B\\\(\quad-\) GPT-OSS-120B\end{tabular} & \(\phantom{{-}0}\llap{\(0\)}.0751\) & \ci{\([\phantom{{-}0}\llap{\(0\)}.0206, 0.1301]\)} & \(0.0279\)\\[3pt]
\bottomrule
\end{tabular}%
\endgroup

\end{table*}

\subsection{Execution checks}

All nine checkpoints completed preflight, smoke, and full execution stages with
accepted atomic-pair, identity-control, and native-dispatch checks. Following a
resource interruption during the Gemma run, 44 complete pairs were retained
from the initial invocation and a resumed invocation supplied the remaining
246 (run IDs are recorded in the archived execution logs). Any unfinished pair
was recomputed in full so that intervention cells were never joined across
invocations.

\begingroup\parfillskip=0pt plus .5\linewidth\finalhyphendemerits=100000
Execution version, scientific source, and configuration remained unchanged, and
the recovery procedure does not alter the receiver-pair estimand or the
source-component resampling unit. Recorded full-run and smoke forwards have a
lower bound of 63,488. Work interrupted before completion is not recorded
exactly, so the fresh-plan count of 63,216 should not be read as an actual
execution total.
\par\endgroup

The secondary verification reconstructed all 54 original mean/total endpoints
and reproduced their stored point estimates, interval endpoints, and bootstrap
standard errors before computing additional contrasts. The manuscript analysis
reuses this accepted primary analysis and introduces no additional model
forwards or protected-validation access. Artifact identities are listed in
Appendix~\ref{app:records}.

\section{Surface Controls and Routing Diagnostics}
\label{app:surface-routing}

The fixed-receiver contrast $D$ holds the receiver constant, but its two
donors differ in both correctness and content. We first ask how much of the
residual contrast surface differences between same-label donors can
reproduce, and then whether router interventions change expert selection,
whether the two donor roles in $D$ perturb routing by similar amounts, and how
large routing effects are relative to residual effects and to $T$. All
analyses reuse the saved outputs and diagnostics of the existing intervention
runs.

\subsection{Surface-oriented same-label placebo}
\label{app:same-label-placebo}

\paragraph{Construction and inference.}
The fixed-receiver and token-window runs use the same donor-ranking rule
with different seeds for breaking ties. For 195 of the 290 receiver pairs,
they selected different external donor pairs. This subset contains 166
receiver source components and is identical across models. For each
receiver trace, we compare its patched margins under the two external
donors with the same correctness label: both erroneous for a correct
receiver, and both correct for an erroneous receiver. The intervention is
at the last-token post-MoE residual interface in both runs.

To orient this placebo, we use the out-of-fold error score of the TF-IDF
text classifier. The placebo is the margin under the higher-scoring donor
minus the margin under the lower-scoring donor. We average the two receiver
roles within each pair, as for $D_{\mathrm{common}}$. Thus, a positive
placebo means that the more error-like surface content produces a more
\texttt{INCORRECT}-favoring judgment even though the donor label is fixed.
We compare this quantity with real $D_{\mathrm{common}}$ on the same
retained receiver pairs; below, $D$ denotes $D_{\mathrm{common}}$.
There are no TF-IDF ties in the eligible subset. Qwen3 is excluded because
its fixed-receiver and token-window runs intervene at different layers.

We use 95\% receiver source-component percentile intervals with 50,000
PCG64 resamples and seed 0. Real $D$, placebo, and their difference share
the same resamples, so the difference interval is paired. This is a post
hoc analysis of existing runs. Before computing the interval estimates,
we fixed three conditions for a checkpoint to pass:
\begin{enumerate}[label=(\alph*),leftmargin=*]
  \item the lower confidence bound of real $D$ is above zero;
  \item the placebo interval includes zero, or its upper bound is below
  one-quarter of the real-$D$ point estimate; and
  \item the lower confidence bound of real $D$ minus placebo is above zero.
\end{enumerate}
The aggregate criterion requires at least six of the eight eligible checkpoints
to pass all three conditions.

\begin{table}[tbp]
\centering
\caption{Same-label placebo at the last-token residual interface.
Real $D$, TF-IDF-oriented placebo, and their paired difference use the
same 195 receiver pairs in 166 source components. Brackets give 95\%
source-component percentile intervals. Differences are computed before rounding.
All checkpoints except Qwen3.6 and
OLMoE meet the three conditions in Appendix~\ref{app:same-label-placebo}.}
\label{tab:same-label-placebo}
\begingroup
\renewcommand{\arraystretch}{1.08}
\tablestyle
\begin{tabular}{@{}lccc@{}}
\toprule
Model & Real $D$ & Placebo & Real $D$ $-$ placebo \\
\midrule
Qwen3.6 & \begin{tabular}[t]{@{}c@{}}\(0.008\)\\\ci{\([{-}0.014, 0.031]\)}\end{tabular} & \begin{tabular}[t]{@{}c@{}}\({-}0.013\)\\\ci{\([{-}0.039, 0.013]\)}\end{tabular} & \begin{tabular}[t]{@{}c@{}}\(0.021\)\\\ci{\([{-}0.016, 0.062]\)}\end{tabular}\\[2pt]
GLM-4.7-Flash & \begin{tabular}[t]{@{}c@{}}\(0.183\)\\\ci{\([\phantom{{-}0}\llap{\(0\)}.144, 0.222]\)}\end{tabular} & \begin{tabular}[t]{@{}c@{}}\({-}0.004\)\\\ci{\([{-}0.036, 0.029]\)}\end{tabular} & \begin{tabular}[t]{@{}c@{}}\(0.186\)\\\ci{\([\phantom{{-}0}\llap{\(0\)}.134, 0.238]\)}\end{tabular}\\[2pt]
Lightning & \begin{tabular}[t]{@{}c@{}}\(0.140\)\\\ci{\([\phantom{{-}0}\llap{\(0\)}.117, 0.164]\)}\end{tabular} & \begin{tabular}[t]{@{}c@{}}\(\phantom{{-}0}\llap{\(0\)}.016\)\\\ci{\([{-}0.003, 0.035]\)}\end{tabular} & \begin{tabular}[t]{@{}c@{}}\(0.124\)\\\ci{\([\phantom{{-}0}\llap{\(0\)}.093, 0.156]\)}\end{tabular}\\[2pt]
Nano & \begin{tabular}[t]{@{}c@{}}\(0.075\)\\\ci{\([\phantom{{-}0}\llap{\(0\)}.063, 0.088]\)}\end{tabular} & \begin{tabular}[t]{@{}c@{}}\(\phantom{{-}0}\llap{\(0\)}.004\)\\\ci{\([{-}0.006, 0.014]\)}\end{tabular} & \begin{tabular}[t]{@{}c@{}}\(0.072\)\\\ci{\([\phantom{{-}0}\llap{\(0\)}.055, 0.088]\)}\end{tabular}\\[2pt]
Gemma 4 & \begin{tabular}[t]{@{}c@{}}\(0.805\)\\\ci{\([\phantom{{-}0}\llap{\(0\)}.558, 1.068]\)}\end{tabular} & \begin{tabular}[t]{@{}c@{}}\({-}0.157\)\\\ci{\([{-}0.418, 0.105]\)}\end{tabular} & \begin{tabular}[t]{@{}c@{}}\(0.961\)\\\ci{\([\phantom{{-}0}\llap{\(0\)}.562, 1.381]\)}\end{tabular}\\[2pt]
OLMoE & \begin{tabular}[t]{@{}c@{}}\(0.169\)\\\ci{\([\phantom{{-}0}\llap{\(0\)}.102, 0.241]\)}\end{tabular} & \begin{tabular}[t]{@{}c@{}}\(\phantom{{-}0}\llap{\(0\)}.111\)\\\ci{\([\phantom{{-}0}\llap{\(0\)}.062, 0.163]\)}\end{tabular} & \begin{tabular}[t]{@{}c@{}}\(0.057\)\\\ci{\([{-}0.009, 0.124]\)}\end{tabular}\\[2pt]
GPT-OSS-20B & \begin{tabular}[t]{@{}c@{}}\(0.130\)\\\ci{\([\phantom{{-}0}\llap{\(0\)}.095, 0.167]\)}\end{tabular} & \begin{tabular}[t]{@{}c@{}}\(\phantom{{-}0}\llap{\(0\)}.030\)\\\ci{\([{-}0.012, 0.070]\)}\end{tabular} & \begin{tabular}[t]{@{}c@{}}\(0.100\)\\\ci{\([\phantom{{-}0}\llap{\(0\)}.051, 0.151]\)}\end{tabular}\\[2pt]
GPT-OSS-120B & \begin{tabular}[t]{@{}c@{}}\(0.213\)\\\ci{\([\phantom{{-}0}\llap{\(0\)}.175, 0.254]\)}\end{tabular} & \begin{tabular}[t]{@{}c@{}}\(\phantom{{-}0}\llap{\(0\)}.019\)\\\ci{\([{-}0.015, 0.054]\)}\end{tabular} & \begin{tabular}[t]{@{}c@{}}\(0.194\)\\\ci{\([\phantom{{-}0}\llap{\(0\)}.144, 0.247]\)}\end{tabular}\\[2pt]
\bottomrule
\end{tabular}
\endgroup

\end{table}

\paragraph{Results and sensitivity.}
Six of eight checkpoints pass (Table~\ref{tab:same-label-placebo}). Qwen3.6
does not have a resolved positive residual $D$, consistent with the
full-sample result. OLMoE has a positive same-label placebo amounting to
about two-thirds of its real $D$. Its real-minus-placebo contrast is
$0.057\;[-0.009,0.124]$, so a substantial part of its residual $D$ is
associated with surface content.

Orienting the placebo by character count gives six passing checkpoints, whereas
word-count orientation and the total-logprob margin each give five.
The count-based orientations subtract the shorter donor's margin from
the longer donor's margin. They drop a receiver pair if either same-label
comparison ties, leaving 194 pairs for character count and 183 for word
count; real $D$ is recomputed on the corresponding subset. The
total-logprob sensitivity retains TF-IDF orientation and all 195 pairs.
GPT-OSS-20B is the additional checkpoint that fails the criterion for word
count and total log probability: its placebo is positive, but real $D$
minus placebo remains positive. Character-count orientation is an
additional sensitivity analysis, separate from the conditions fixed
before estimation.

The category-exclusion sensitivity removes circular and redundancy pairs,
leaving 281 pairs overall and 194 placebo-eligible pairs in 166 source
components. The TF-IDF-oriented test still passes in six of eight checkpoints
with the same two exceptions (bootstrap seed 20260908).

This comparison has two limitations. First, eligibility depends on ties
in the donor-ranking rule, so these donors and receiver pairs are not a
random sample for the placebo test. Second, the two margins come from
different runs. For Nano and Lightning, even identical donors produce
nonzero cross-run differences, so their placebo includes numerical drift.
Both still pass when each patched margin is first differenced from its
own run's unpatched receiver margin. For the models that pass, the supported
interpretation is dependence on donor correctness beyond this lexical proxy;
the test does not isolate pure correctness or rule out surface features that TF-IDF
does not capture.

\subsection{Residual effects when the donor surface gap is small}
\label{app:surface-gap}

The placebo compares different donors with the same label. A complementary
analysis asks whether real $D$ persists when the TF-IDF classifier does
not rank the erroneous donor as more error-like than the correct donor.
For the external donor pair used in $D$, define
$\Delta s=s(\error_j)-s(\clean_j)$, where $s$ is the out-of-fold TF-IDF
error score. We report the average residual $D$ on pairs with
$\Delta s\leq 0$. We also fit an ordinary least-squares regression of
pair-level $D$ on $\Delta s$ using all 290 receiver pairs and report its
intercept at zero gap. The regressions are refitted within each of 50,000
source-component resamples with seed 0. Qwen3 is excluded because its
archived out-of-fold scores do not align with the evaluated boundary
steps.

\begin{table}[tbp]
\centering
\caption{Residual $D$ and the TF-IDF donor gap.
The left column uses the 80 receiver pairs in 78 source components for
which the erroneous donor's TF-IDF error score is no higher than the
correct donor's. The right column is the zero-gap intercept of a linear
regression on all 290 pairs in 235 source components. These are distinct
estimands. Brackets give 95\% source-component percentile intervals.}
\label{tab:surface-gap}
\begingroup
\renewcommand{\arraystretch}{1.08}
\tablestyle
\begin{tabular}{@{}lcc@{}}
\toprule
Model & $D$ where $\Delta s\leq 0$ & Intercept at $\Delta s=0$ \\
\midrule
Qwen3.6 & \begin{tabular}[t]{@{}c@{}}\(0.038\)\\\ci{\([\phantom{{-}0}\llap{\(0\)}.003, 0.081]\)}\end{tabular} & \begin{tabular}[t]{@{}c@{}}\(0.021\)\\\ci{\([{-}0.002, 0.047]\)}\end{tabular}\\[2pt]
GLM-4.7-Flash & \begin{tabular}[t]{@{}c@{}}\(0.124\)\\\ci{\([\phantom{{-}0}\llap{\(0\)}.071, 0.182]\)}\end{tabular} & \begin{tabular}[t]{@{}c@{}}\(0.145\)\\\ci{\([\phantom{{-}0}\llap{\(0\)}.109, 0.182]\)}\end{tabular}\\[2pt]
Lightning & \begin{tabular}[t]{@{}c@{}}\(0.104\)\\\ci{\([\phantom{{-}0}\llap{\(0\)}.073, 0.137]\)}\end{tabular} & \begin{tabular}[t]{@{}c@{}}\(0.125\)\\\ci{\([\phantom{{-}0}\llap{\(0\)}.101, 0.149]\)}\end{tabular}\\[2pt]
Nano & \begin{tabular}[t]{@{}c@{}}\(0.062\)\\\ci{\([\phantom{{-}0}\llap{\(0\)}.045, 0.081]\)}\end{tabular} & \begin{tabular}[t]{@{}c@{}}\(0.065\)\\\ci{\([\phantom{{-}0}\llap{\(0\)}.052, 0.078]\)}\end{tabular}\\[2pt]
Gemma 4 & \begin{tabular}[t]{@{}c@{}}\(1.024\)\\\ci{\([\phantom{{-}0}\llap{\(0\)}.602, 1.478]\)}\end{tabular} & \begin{tabular}[t]{@{}c@{}}\(0.864\)\\\ci{\([\phantom{{-}0}\llap{\(0\)}.614, 1.130]\)}\end{tabular}\\[2pt]
OLMoE & \begin{tabular}[t]{@{}c@{}}\(0.001\)\\\ci{\([{-}0.022, 0.024]\)}\end{tabular} & \begin{tabular}[t]{@{}c@{}}\(0.056\)\\\ci{\([\phantom{{-}0}\llap{\(0\)}.012, 0.102]\)}\end{tabular}\\[2pt]
GPT-OSS-20B & \begin{tabular}[t]{@{}c@{}}\(0.060\)\\\ci{\([\phantom{{-}0}\llap{\(0\)}.027, 0.095]\)}\end{tabular} & \begin{tabular}[t]{@{}c@{}}\(0.076\)\\\ci{\([\phantom{{-}0}\llap{\(0\)}.046, 0.106]\)}\end{tabular}\\[2pt]
GPT-OSS-120B & \begin{tabular}[t]{@{}c@{}}\(0.151\)\\\ci{\([\phantom{{-}0}\llap{\(0\)}.095, 0.215]\)}\end{tabular} & \begin{tabular}[t]{@{}c@{}}\(0.158\)\\\ci{\([\phantom{{-}0}\llap{\(0\)}.123, 0.195]\)}\end{tabular}\\[2pt]
\bottomrule
\end{tabular}
\endgroup

\end{table}

There are 80 pairs with $\Delta s\leq 0$. Residual $D$ remains positive
in seven of eight checkpoints on this subset (Table~\ref{tab:surface-gap}),
including Qwen3.6, whose full-sample contrast is unresolved. OLMoE is the
exception: its subset estimate is $0.001\;[-0.022,0.024]$. Its fitted
zero-gap intercept is nevertheless positive,
$0.056\;[0.012,0.102]$. The intercept uses a linear fit across the full
gap range, whereas the subset mean averages only observations at or below
zero. Together with the placebo, these results support a substantial
surface-associated component in OLMoE without establishing that all of its
residual effect is explained by TF-IDF. The gap analysis is correlational.

\subsection{Routing manipulation and donor perturbation size}
\label{app:routing-manipulation}

\paragraph{Measurements.}
For each router intervention, the saved diagnostics record whether any
member of the selected top-$k$ expert set changes, the fraction of experts
retained, and the KL divergence from receiver to replacement router
logits after applying softmax over all experts. A changed top-$k$ set
means that at least one selected expert differs; this measure does not
detect weight changes within an unchanged set. We average each statistic
over the two candidate-scoring forwards and over patched router sites
within a receiver trace, then average across receiver traces. The last-token
single-layer comparisons have 580 receiver traces per model and treatment.

\begin{table}[tbp]
\centering
\caption{Routing changes and perturbation-size comparisons.
Top-$k$ changed is the mean changed-set fraction for a paired last-token
single-layer router intervention. The matched/paired KL range spans all
four token windows and both single-layer and all-router interventions.
The last two columns compare the erroneous and correct external donors
used in $D$: the ratio of their mean KL values at the last-token single
layer and under full-step all-router replacement, respectively.}
\label{tab:routing-manipulation}
\begingroup
\renewcommand{\arraystretch}{1.08}
\tablestyle
\begin{tabular}{@{}lcccc@{}}
\toprule
Model & \shortstack{Top-$k$\\changed} & \shortstack{Matched/paired\\KL range} & \shortstack{$D$ donors: E/C KL\\single layer} & \shortstack{$D$ donors: E/C KL\\all-router} \\
\midrule
Qwen3.6 & 0.926 & 1.715--2.119 & 0.998 & 0.998 \\[2pt]
Qwen3 & 0.683 & 1.585--2.144 & 0.939 & 0.995 \\[2pt]
GLM-4.7-Flash & 0.636 & 1.657--2.331 & 0.999 & 1.016 \\[2pt]
Lightning & 0.883 & 1.763--2.160 & 0.944 & 1.010 \\[2pt]
Nano & 0.843 & 1.738--2.141 & 0.900 & 1.009 \\[2pt]
Gemma 4 & 0.886 & 1.560--1.934 & 1.091 & 1.009 \\[2pt]
OLMoE & 0.348 & 1.612--2.213 & 1.030 & 1.024 \\[2pt]
GPT-OSS-20B & 0.655 & 1.676--2.224 & 1.049 & 0.999 \\[2pt]
GPT-OSS-120B & 0.709 & 1.636--2.202 & 0.955 & 1.000 \\[2pt]
\bottomrule
\end{tabular}
\endgroup

\end{table}

Single-layer paired router interventions change top-$k$ membership in
approximately 35\% to 93\% of cases across models
(Table~\ref{tab:routing-manipulation}). In the fixed-receiver runs used
for $D$, the corresponding changed-set fraction ranges from approximately
45\% to 100\% across models and donor roles. Thus, unresolved single-layer
router $D$ does not imply that expert selection is unchanged. Across the
four windows and the two router intervention scopes, matched external
donors produce approximately 1.6 to 2.3 times the mean router KL of paired
donors. In contrast, the erroneous-to-correct donor KL ratio in the
fixed-receiver $D$ runs ranges from 0.90 to 1.09 across the single-layer
and full-step all-router settings. The two donor roles therefore produce
similar average perturbation sizes, although this comparison does not
establish that individual interventions are matched in every respect.

\paragraph{Interpretation boundaries.}
KL here is a shared projection of router logits, not necessarily the
divergence of the native dispatch weights. Some models use gate functions
other than softmax over all experts, and raw KL values should not be
compared across models. Under all-router replacement, the receiver logits
at a later layer are computed after earlier layers have already been
patched; the diagnostic is relative to that forward's state, not the
native unpatched routing. At multi-token windows, paired donors also
share prefix positions with the receiver, whereas matched donors do not.
Part of their perturbation-size difference is therefore positional.

\subsection{Routing effects relative to the residual effect and to $T$}
\label{app:relative-effects}

Table~\ref{tab:relative-effects} expresses the full-step interventions on a
relative scale. The first three columns give the paired effect $T$ of the
single-layer router, single-layer MoE-output, and all-layer router
interventions as a percentage of the single-layer residual $T$ at the same
token window. Each entry is a ratio of means over the same pairs, with a
95\% percentile-bootstrap interval in which numerator and denominator share
resamples. Over the full step, the single-layer router effect is at most
5.9\% of the residual effect, whereas the all-layer router effect ranges
from 24.8\% to 133.8\%.

The last column divides the full-step all-router $D_{\mathrm{common}}$ of
Table~\ref{tab:wide-d-primary}, and the two ends of its interval, by the
full-step all-router $T$. If donors with the receiver's own label left the
margin unchanged and matched donors acted like paired donors,
$D_{\mathrm{common}}$ would equal $T$ under Equations~\ref{eq:t} and~\ref{eq:d}. The
ratio can therefore be read as the share of the routing effect that depends
on donor correctness. Its upper interval end is below 7\% in four
checkpoints (Qwen3, GLM-4.7-Flash, Nano, and GPT-OSS-120B) and below 15\% in
two more (Lightning and Qwen3.6). In Gemma and OLMoE the interval is too
wide to bound the share, and GPT-OSS-20B is the only checkpoint with a
resolved positive share.

\begin{table}[tbp]
\centering
\caption{Routing effects on a relative scale (full step).
Columns 2 to 4: paired effect $T$ as a percentage of the single-layer
residual $T$, with 95\% intervals. Column 5: all-router $D_{\mathrm{common}}$
and its interval ends as a percentage of the all-router $T$ point estimate.}
\label{tab:relative-effects}
\begingroup
\renewcommand{\arraystretch}{1.08}
\tablestyle
\begin{tabular}{@{}lcccc@{}}
\toprule
 & \multicolumn{3}{c}{$T$ as \% of single-layer residual $T$} & \shortstack{All-router $D_{\mathrm{common}}$\\as \% of all-router $T$} \\
\cmidrule(lr){2-4}
Model & \shortstack{Router,\\one layer} & \shortstack{MoE output,\\one layer} & \shortstack{Router,\\all layers} & \\
\midrule
Qwen3.6 & \({-}0.2\)\hspace{0.25em}\ci{\([{-}1.0, 0.6]\)} & \(\phantom{23}\llap{\(0\)}.3\)\hspace{0.25em}\ci{\([{-}0.8, \phantom{26}\llap{\(1\)}.3]\)} & \(\phantom{101}\llap{\(27\)}.0\)\hspace{0.25em}\ci{\([\phantom{112}\llap{\(23\)}.2, \phantom{114}\llap{\(30\)}.9]\)} & \(\phantom{{-}5}\llap{\(3\)}.0\)\hspace{0.25em}\ci{\([\phantom{{-}11}\llap{\({-}8\)}.1, 14.0]\)} \\[2pt]
Qwen3 & \(\phantom{{-}0}\llap{\(0\)}.5\)\hspace{0.25em}\ci{\([\phantom{{-}1}\llap{\(0\)}.0, 1.1]\)} & \(\phantom{23}\llap{\(2\)}.8\)\hspace{0.25em}\ci{\([\phantom{{-}0}\llap{\(1\)}.9, \phantom{26}\llap{\(3\)}.8]\)} & \(\phantom{101}\llap{\(37\)}.5\)\hspace{0.25em}\ci{\([\phantom{112}\llap{\(33\)}.7, \phantom{114}\llap{\(41\)}.2]\)} & \({-}5.4\)\hspace{0.25em}\ci{\([{-}11.9, \phantom{14}\llap{\(1\)}.1]\)} \\[2pt]
GLM-4.7-Flash & \(\phantom{{-}0}\llap{\(1\)}.4\)\hspace{0.25em}\ci{\([{-}0.6, 3.3]\)} & \(23.3\)\hspace{0.25em}\ci{\([\phantom{{-}0}\llap{\(20\)}.3, 26.3]\)} & \(\phantom{101}\llap{\(54\)}.8\)\hspace{0.25em}\ci{\([\phantom{112}\llap{\(47\)}.1, \phantom{114}\llap{\(62\)}.7]\)} & \({-}5.4\)\hspace{0.25em}\ci{\([{-}15.7, \phantom{14}\llap{\(4\)}.8]\)} \\[2pt]
Lightning & \(\phantom{{-}0}\llap{\(0\)}.1\)\hspace{0.25em}\ci{\([{-}1.6, 1.6]\)} & \(\phantom{23}\llap{\(7\)}.1\)\hspace{0.25em}\ci{\([\phantom{{-}0}\llap{\(4\)}.9, \phantom{26}\llap{\(9\)}.3]\)} & \(101.5\)\hspace{0.25em}\ci{\([\phantom{112}\llap{\(88\)}.7, 114.9]\)} & \(\phantom{{-}5}\llap{\(5\)}.8\)\hspace{0.25em}\ci{\([\phantom{{-}11}\llap{\({-}2\)}.3, 13.8]\)} \\[2pt]
Nano & \({-}0.4\)\hspace{0.25em}\ci{\([{-}2.6, 1.8]\)} & \(10.3\)\hspace{0.25em}\ci{\([\phantom{{-}0}\llap{\(7\)}.9, 12.7]\)} & \(110.4\)\hspace{0.25em}\ci{\([\phantom{112}\llap{\(95\)}.9, 125.4]\)} & \({-}2.2\)\hspace{0.25em}\ci{\([\phantom{{-}11}\llap{\({-}9\)}.3, \phantom{14}\llap{\(4\)}.8]\)} \\[2pt]
Gemma 4 & \(\phantom{{-}0}\llap{\(1\)}.2\)\hspace{0.25em}\ci{\([\phantom{{-}1}\llap{\(0\)}.4, 2.1]\)} & \(18.5\)\hspace{0.25em}\ci{\([\phantom{{-}0}\llap{\(15\)}.8, 21.3]\)} & \(\phantom{101}\llap{\(24\)}.8\)\hspace{0.25em}\ci{\([\phantom{112}\llap{\(20\)}.0, \phantom{114}\llap{\(29\)}.8]\)} & \(\phantom{{-}5}\llap{\(18\)}.3\)\hspace{0.25em}\ci{\([\phantom{{-}11}\llap{\({-}0\)}.9, 37.6]\)} \\[2pt]
OLMoE & \(\phantom{{-}0}\llap{\(3\)}.1\)\hspace{0.25em}\ci{\([{-}1.6, 7.8]\)} & \(19.2\)\hspace{0.25em}\ci{\([\phantom{{-}0}\llap{\(12\)}.5, 26.0]\)} & \(\phantom{101}\llap{\(57\)}.1\)\hspace{0.25em}\ci{\([\phantom{112}\llap{\(41\)}.5, \phantom{114}\llap{\(73\)}.8]\)} & \(\phantom{{-}5}\llap{\(8\)}.8\)\hspace{0.25em}\ci{\([{-}20.3, 38.5]\)} \\[2pt]
GPT-OSS-20B & \(\phantom{{-}0}\llap{\(1\)}.4\)\hspace{0.25em}\ci{\([\phantom{{-}1}\llap{\(0\)}.5, 2.4]\)} & \(\phantom{23}\llap{\(9\)}.7\)\hspace{0.25em}\ci{\([\phantom{{-}0}\llap{\(7\)}.9, 11.4]\)} & \(\phantom{101}\llap{\(25\)}.4\)\hspace{0.25em}\ci{\([\phantom{112}\llap{\(21\)}.4, \phantom{114}\llap{\(29\)}.5]\)} & \(\phantom{{-}5}\llap{\(23\)}.6\)\hspace{0.25em}\ci{\([\phantom{{-}11}\llap{\(6\)}.1, 41.3]\)} \\[2pt]
GPT-OSS-120B & \(\phantom{{-}0}\llap{\(5\)}.9\)\hspace{0.25em}\ci{\([\phantom{{-}1}\llap{\(2\)}.4, 9.4]\)} & \(15.4\)\hspace{0.25em}\ci{\([\phantom{{-}0}\llap{\(11\)}.8, 18.8]\)} & \(133.8\)\hspace{0.25em}\ci{\([112.9, 156.3]\)} & \({-}1.2\)\hspace{0.25em}\ci{\([\phantom{{-}11}\llap{\({-}9\)}.2, \phantom{14}\llap{\(6\)}.8]\)} \\[2pt]
\bottomrule
\end{tabular}
\endgroup

\end{table}

\paragraph{Interpretation boundaries.}
These ratios are descriptive and were computed after the main analyses. The
denominator of the last column is a point estimate, so its uncertainty is
not propagated. $T$ uses paired donors, whereas $D$ uses matched donors from
another problem. No equivalence margin was specified in advance, and an
interval containing zero does not establish equivalence.

\section{Natural-Error Evaluation on ProcessBench}
\label{app:processbench-details}

The single-layer residual-versus-router difference also appears on naturally
occurring errors. With three checkpoints and 64 pairs, and without fixed-receiver
$D$ or token-window analyses, this is an exploratory extension, kept separate
from the nine-checkpoint study.

\subsection{Evaluation setup}

From the public GSM8K subset of ProcessBench \citep{zheng2025processbench}, 193
of 400 source traces are all-correct and 207 contain a first-error label. We
construct 64 same-generator pairs, each linking a trace with a naturally
occurring first error to an all-correct trace at the same step index. The
resulting 128 traces contain no trace reuse, and all 64 pairs are
cross-problem matches. GPT-OSS-20B, Gemma, and Nemotron-3 Nano use the same
processed-data selection, binary verification prompt, boundary policy, two
replacement directions, $\alpha=1$, identity conditions, and outcome metric as
the main study, with unchanged architecture-specific anchors. A scoped Nemotron
adapter repair changed neither the samples, boundaries, estimand, nor
intervention comparisons. On the balanced 128-trace endpoint, GPT-OSS-20B has
accuracy 0.625 and AUROC 0.736, Gemma 0.688 and 0.854, and Nano 0.641 and
0.824. Each full run contains 128 behavior rows, 128 recapture rows, and 768
intervention rows, and artifact-integrity and coverage checks are independent
of effect direction and statistical significance.

\subsection{Interface comparison}

Residual interventions produce larger absolute changes than router-logit
interventions in all six checkpoint-by-direction comparisons, and the paired
residual-minus-router $T$ difference is positive in all three checkpoints
(Table~\ref{tab:processbench}). Residual directional intervals exclude zero in
four of the six comparisons, whereas router intervals exclude zero in none.
Five of six residual point estimates move in the direction associated with the
donor's correctness. Nano has the expected sign in both directions, GPT-OSS is
clear only in the correct-to-error direction, and Gemma has a large expected
effect in the reverse direction but not in the correct-to-error direction,
which motivates the receiver-state analysis below.

\begin{table*}[tbp]
\centering
\renewcommand{\arraystretch}{1.0}
\caption{ProcessBench natural-error transfer on 64 same-generator,
cross-problem pairs per model. MC$\rightarrow$FE uses a matched-correct donor and
first-error receiver; FE$\rightarrow$MC swaps their roles. Each cell shows an
estimate above its paired-cluster percentile-bootstrap 95\% interval (50,000
resamples; no multiplicity adjustment). \(T\) and \(G\) come directly from paired
rows, not marginal intervals.}
\label{tab:processbench}
\tablestyle
\begin{tabular}{llcccc}
\toprule
Model & Site & MC$\rightarrow$FE & FE$\rightarrow$MC & \(T\) & \(G=T_{\resid}-T_{\router}\) \\
\midrule
\multirow{4}{*}{GPT-OSS 20B} & residual & \({-}.0590\) & \({+}.0242\) & \(\phantom{{-}}.0416\) & \multirow{4}{*}{\begin{tabular}[c]{@{}c@{}}\(.0476\)\\\ci{\([.0145, .0817]\)}\end{tabular}} \\
& & \ci{\([{-}.1029, {-}.0163]\)} & \ci{\([{-}.0344, .0824]\)} & \ci{\([\phantom{{-}}.0065, .0775]\)} & \\
& router & \({+}.0051\) & \({-}.0070\) & \({-}.0060\) & \\
& & \ci{\([{-}.0060, \phantom{{-}}.0164]\)} & \ci{\([{-}.0174, .0035]\)} & \ci{\([{-}.0143, .0025]\)} & \\
\addlinespace[2pt]
\multirow{4}{*}{Gemma 4 26B-A4B} & residual & \({+}.0442\) & \({+}.2948\) & \(\phantom{{-}}.1253\) & \multirow{4}{*}{\begin{tabular}[c]{@{}c@{}}\(.1241\)\\\ci{\([.0246, .2402]\)}\end{tabular}} \\
& & \ci{\([{-}.0813, \phantom{{-}}.1594]\)} & \ci{\([\phantom{{-}}.1424, .5035]\)} & \ci{\([\phantom{{-}}.0224, .2452]\)} & \\
& router & \({+}.0010\) & \({+}.0034\) & \(\phantom{{-}}.0012\) & \\
& & \ci{\([{-}.0339, \phantom{{-}}.0358]\)} & \ci{\([{-}.0203, .0268]\)} & \ci{\([{-}.0183, .0205]\)} & \\
\addlinespace[2pt]
\multirow{4}{*}{Nemotron-3 Nano} & residual & \({-}.0517\) & \({+}.0667\) & \(\phantom{{-}}.0592\) & \multirow{4}{*}{\begin{tabular}[c]{@{}c@{}}\(.0575\)\\\ci{\([.0396, .0755]\)}\end{tabular}} \\
& & \ci{\([{-}.0686, {-}.0345]\)} & \ci{\([\phantom{{-}}.0454, .0897]\)} & \ci{\([\phantom{{-}}.0410, .0778]\)} & \\
& router & \({-}.0033\) & \({+}.0001\) & \(\phantom{{-}}.0017\) & \\
& & \ci{\([{-}.0093, \phantom{{-}}.0027]\)} & \ci{\([{-}.0054, .0056]\)} & \ci{\([{-}.0024, .0056]\)} & \\
\bottomrule
\end{tabular}
\end{table*}

\subsection{Receiver-state analysis}
\label{app:processbench-explanation}

We reanalyze the same 64 pairs in all three checkpoints, retaining both
replacement directions and both intervention sites, and group receivers by
whether their unpatched verification margin correctly judges the evaluated
boundary. Group intervals condition on the observed group sizes: 50,000 PCG64
draws with seed 0 resample pairs independently within the two groups, and group
differences are computed from the joint draws. These exploratory, unadjusted
intervals describe an observational grouping. Because the baseline margin that
defines the groups also enters the intervention effect with a negative sign,
the group comparisons are subject to baseline conditioning.

\begin{figure}[tbp]
\centering
\includegraphics[width=\textwidth]{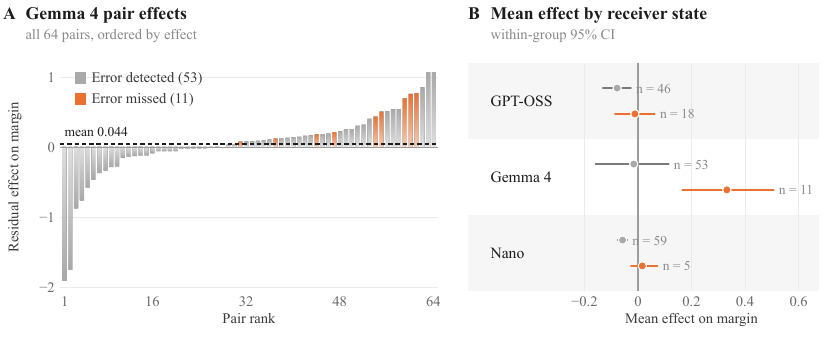}
\caption{Exploratory correct-to-error residual intervention on ProcessBench.
\textbf{A}: all 64 Gemma pair effects, ordered by value; the dashed line is their
mean. \textbf{B}: mean effects grouped by whether the receiver initially detects
the error, with group sizes $n$. Intervals use 50,000 percentile bootstrap
draws within the observed receiver-state groups. Negative effects correspond to
the expected direction for this intervention.}
\label{fig:processbench-receiver-state}
\end{figure}

\begingroup\parfillskip=0pt plus .5\linewidth\finalhyphendemerits=100000
Gemma's positive correct-to-error point estimate is not attributable to a single
pair (Figure~\ref{fig:processbench-receiver-state}): 37 of 64 effects are
positive, the median is $0.078$, and all 64 leave-one-out means lie between
$0.028$ and $0.075$. The mean is $0.332$ for the 11 initially missed errors and
$-0.016$ for the 53 initially detected errors, a difference of
$0.348\;[0.134,0.571]$. Not every positive effect belongs to the missed-error
group, however: 28 of 53 detected-error receivers also have positive effects.
Table~\ref{tab:processbench-receiver-state} reports the full two-direction
breakdown.
\par\endgroup

\begin{table}[tbp]
\centering
\renewcommand{\arraystretch}{1.12}
\caption{Exploratory residual-transfer effects by the receiver's unpatched
judgment. Entries give mean shifts in nats/target-token; the group contrast is
misjudged minus correctly judged, with marginal 95\% intervals from 50,000
within-group pair-bootstrap draws. C$\rightarrow$E is matched correct to first
error; E$\rightarrow$C is the reverse. All original pairs remain included.}
\label{tab:processbench-receiver-state}
\tablestyle
\begin{tabular}{@{}lcrrc@{}}
\toprule
Model & Direction & \shortstack{Correctly judged\\$n$, mean} &
\shortstack{Misjudged\\$n$, mean} & \shortstack{Group contrast\\$[95\%\ \mathrm{CI}]$} \\
\midrule
GPT-OSS & C$\rightarrow$E & 46, \({-}.0776\) & 18, \({-}.0114\) & \(\phantom{{-}}.0662\)\hspace{0.25em}\ci{\([{-}.0249, \phantom{{-}}.1564]\)} \\
        & E$\rightarrow$C & 34, \(\phantom{{-}}.0436\) & 30, \(\phantom{{-}}.0022\) & \({-}.0415\)\hspace{0.25em}\ci{\([{-}.1618, \phantom{{-}}.0747]\)} \\
\addlinespace[3pt]
Gemma 4 & C$\rightarrow$E & 53, \({-}.0155\) & 11, \(\phantom{{-}}.3320\) & \(\phantom{{-}}.3475\)\hspace{0.25em}\ci{\([\phantom{{-}}.1342, \phantom{{-}}.5708]\)} \\
        & E$\rightarrow$C & 35, \(\phantom{{-}}.3126\) & 29, \(\phantom{{-}}.2733\) & \({-}.0392\)\hspace{0.25em}\ci{\([{-}.4320, \phantom{{-}}.2568]\)} \\
\addlinespace[3pt]
Nano    & C$\rightarrow$E & 59, \({-}.0574\) & 5, \(\phantom{{-}}.0164\) & \(\phantom{{-}}.0738\)\hspace{0.25em}\ci{\([\phantom{{-}}.0272, \phantom{{-}}.1340]\)} \\
        & E$\rightarrow$C & 23, \(\phantom{{-}}.1018\) & 41, \(\phantom{{-}}.0469\) & \({-}.0549\)\hspace{0.25em}\ci{\([{-}.1085, {-}.0053]\)} \\
\bottomrule
\end{tabular}
\end{table}

\paragraph{Construction and shared-problem sensitivity.}
The 64 pairs cover 12 generators and first-error step indices zero through five,
with counts 8, 17, 18, 12, 5, and 4. Although 49 pairs have equal full-trace
step counts, the median absolute difference in visible step-text length is 176
characters, so matching the step index does not match the amount of visible
text. The 128 traces represent 125 normalized problems, linking the pairs into
61 connected components: 59 singletons, one two-pair component, and one
three-pair component. A sensitivity analysis that jointly resamples these
components (50,000 draws, seed 0) gives residual-minus-router gaps of
$0.048\;[0.013,0.083]$ for GPT-OSS, $0.124\;[0.024,0.241]$ for Gemma, and
$0.058\;[0.039,0.076]$ for Nano, and a Gemma receiver-state contrast of
$0.348\;[0.130,0.578]$. These supplementary intervals preserve the
positive-gap result without replacing the pair-bootstrap analysis.

\paragraph{Routing changes and signed averages.}
Each checkpoint has 256 of 256 verified dispatch rows, and cross-example router
interventions change native top-$k$ selection in 49, 63, and 61 of 64 rows per
direction for GPT-OSS, Gemma, and Nano. Gemma's near-zero mean router effect
nonetheless averages over pair-level changes with opposing signs: in the
correct-to-error direction, 30 pair effects are positive, 33 negative, and one
zero, with mean absolute magnitude $0.111$ against a signed mean of $0.001$ (the
residual mean absolute magnitude is $0.320$). A small signed router average
therefore does not imply that every individual router intervention leaves the
judgment unchanged. Native top-$k$ changes and output magnitudes serve only as
descriptive manipulation checks, not calibrated causal doses or
mediation quantities.

\section{Reproducibility and Artifact Records}
\label{app:records}

\textls[-10]{The nine-checkpoint inputs come from three campaign archives:
\texttt{campaign\_20260907\_A} (Qwen3.6, Qwen3, GLM),
\texttt{campaign\_20260907\_B} (Lightning, Nano, Gemma, OLMoE), and
\texttt{campaign\_20260907\_C} (GPT-OSS 20B and 120B). Their resolved
configurations, executed adapters, scores, donor manifests, token geometry, and
native analyses define the experimental identities.}\looseness=-1

The \texttt{abc\_synthesis\_20260908} CPU analysis reads these artifacts and
the saved model outputs and computes the shared-window and direct-contrast
analyses. It executes no new model forwards, does not alter the discovery
selection, and does not use the protected validation split.

SHA-256 checksums for the versioned window JSON and the $D$/bundle headline CSV
are stored in the repository release manifests, and machine-readable rows
record the source file, analysis version, and JSON pointer or pair identity.

\paragraph{Full-step all-router $D$ records.}
\begingroup
\raggedright
The separate \texttt{campaign\_20260908\_D/final\_v1} archive retains each
checkpoint's atomic four-cell scores, full input identities, resolved
configuration, adapter and runner hashes, native 50,000-draw analysis, and
recovery records, together with the SHA-256 checksum for its
\texttt{wide\_d\_nine\_model\_v1.json} summary.

The \texttt{wide\_d\_synthesis\_20260908} analysis stores the main
summary and a separately identified
\texttt{wide\_d\_posthoc\_secondary\_contrasts\_v1} artifact. Its receipt
records source-member hashes, the complete GPT input-identity projection,
Gemma candidate-weight checks, and exact reproduction of the original
endpoints. These records do not overwrite the three campaign archives above or
the representation-probe inputs. No protected-validation split was used in the
full-step $D$ experiment or its secondary analysis.
\par
\endgroup

\begingroup\parfillskip=0pt plus .5\linewidth\finalhyphendemerits=100000
Representation-probe and ProcessBench analyses have separate input and analysis
records. Exact model revisions, pair manifests, fold assignments, seeds, and
source hashes are stored in the analysis archives.
\par\endgroup

\section{Scope and Interpretational Notes}
\label{app:claim-boundaries}

\looseness=-1
Across nine diverse open-weight MoE checkpoints, our experiments establish the
four linked results of the main text. First, linear probes on router logits
separate correct from erroneous steps in all eight checkpoints with recorded
activations, outperforming a text-only baseline (Appendix~\ref{app:access}).
Second, while single-layer residual replacements yield larger $T$ than
single-layer router or MoE-output replacements in all 36 checkpoint-by-window
comparisons, all-layer router replacements shift the verification margin beyond
norm-matched random controls in all nine checkpoints and can exceed
single-layer residual replacement, with all-layer MoE-output replacement
likewise exceeding random controls. Third, holding the receiver fixed, router
$D$ is unresolved at a single layer in all nine checkpoints and reliably
positive under all-layer routing only in GPT-OSS-20B (with Gemma positive on
correct receivers alone, and the 95\% upper bound of $D_{\mathrm{common}}$
below 15\% of all-router $T$ in six checkpoints). Fourth, the same
fixed-receiver contrast detects positive correctness dependence at a single
layer in eight of nine checkpoints for residual states (all except Qwen3.6,
with OLMoE partially associated with surface content;
Appendix~\ref{app:surface-routing}) and in five of nine for MoE outputs.

\looseness=-1
These findings characterize interface-level interventions under the tested
layer coverage, token windows, and verification margin, with the
representation-probe and ProcessBench analyses providing complementary evidence
under their own protocols. In particular, our results show that routing is
causally active ($T>0$) yet rarely transfers donor correctness into the
judgment, with unresolved $D$ intervals placing tight upper bounds on the
correctness-dependent share rather than asserting an exact zero. Localizing
these effects to individual experts, formalizing mediation pathways, and
extending the protocol to additional MoE families and non-lexical surface
controls are natural directions for future work.

\end{document}